\documentclass{article} % For LaTeX2e
\usepackage{main,times}

\usepackage{amsmath,amsfonts,bm}

\def\eqref#1{equation~\ref{#1}}
\def\1{\bm{1}}

\DeclareMathAlphabet{\mathsfit}{\encodingdefault}{\sfdefault}{m}{sl}
\SetMathAlphabet{\mathsfit}{bold}{\encodingdefault}{\sfdefault}{bx}{n}

\usepackage{hyperref}
\usepackage{url}
\usepackage[utf8]{inputenc} % allow utf-8 input
\usepackage[T1]{fontenc}    % use 8-bit T1 fonts
\usepackage{hyperref}       % hyperlinks
\usepackage{url}            % simple URL typesetting
\usepackage{booktabs}       % professional-quality tables
\usepackage{amsfonts}       % blackboard math symbols
\usepackage{nicefrac}       % compact symbols for 1/2, etc.
\usepackage{microtype}      % microtypography
\usepackage{amsmath}
\usepackage{xcolor}
\usepackage[normalem]{ulem}
\usepackage{graphicx}
\usepackage{subcaption}

\usepackage{makecell}
\usepackage{adjustbox}
\usepackage{multirow}
\usepackage{float}
\usepackage{amssymb}
\usepackage{pifont}

\newcommand{\cmark}{\ding{51}}
\newcommand{\xmark}{\ding{56}}

\newcommand{\cutsectiondown}{\vspace*{-0.12in}}

\newcommand{\cutparagraphup}{\vspace*{-0.1in}}

\title{Learning to Compose: Improving Object Centric Learning by Injecting Compositionality}

\author{Whie Jung, Jaehoon Yoo, Sungjin Ahn, Seunghoon Hong \\
School of Computing, KAIST \\
\texttt{\{whieya, wogns98, sungjin.ahn, seunghoon.hong\}@kaist.ac.kr}
}

\iclrfinalcopy % Uncomment for camera-ready version, but NOT for submission.
\begin{document}

\maketitle

\begin{abstract}
Learning compositional representation is a key aspect of object-centric learning as it enables flexible systematic generalization and supports complex visual reasoning. 
% However, most of the existing approaches in object centric learning only optimizes auto-encoding objective without considering compositionality. 
However, most of the existing approaches rely on auto-encoding objective, while the compositionality is implicitly imposed by the architectural or algorithmic bias in the encoder. 
This misalignment between auto-encoding objective and learning compositionality often results in failure of capturing meaningful object representations. 
In this study, we propose a novel objective that explicitly encourages compositionality of the representations. 
Built upon the existing object-centric learning framework (\textit{e.g.}, slot attention), our method incorporates additional constraints that an arbitrary mixture of object representations from two images should be valid by maximizing the likelihood of the composite data. 
We demonstrate that incorporating our objective to the existing framework consistently improves the objective-centric learning and enhances the robustness to the architectural choices. 
Codes are available at 
{\footnotesize\url{https://github.com/whieya/Learning-to-compose}}.

% \href{https://github.com/whieya/Learning-to-compose}{https://github.com/whieya/Learning-to-compose}.
% We demonstrate the effectiveness of our method on four datasets and identify superious object-centric representation compared to the reconstruction-only counterparts, thereby highlighting the importance of directly injecting compositionality in object-centric learning. 

% Discovering composable abstraction is a key to in object centric learning, which enables systematic generalization and supporting complex visual reasoning. However, most of the object centric learning approaches exploits only auto-encoding objective without guaranteeing such property. Due to the misalignment between auto-encoding objective and learning compositionality, they often fail to capture meaningful object representations. In this work, we introduce a novel objective that explicitly encourages compositionality of the representations. To this end, we first generate composite data by randomly composing slot representations extracted from two images and our proposed loss maximizes the validity of the composite data. We demonstrate the efficacy of our proposed objective with various object centric tasks including unsupervised object discovery and property prediction tasks.
\end{abstract}

\section{Introduction}
As the world is highly compositional in nature, relatively few composable units, such as objects or words, can describe infinitely many observations. 
Consequently, human intelligence has evolved to recognize the environment as a combination of composable units, (\textit{e.g.,} objects) which enables rapid adaptation to unseen situations by recomposing the already learned concepts~\citep{spelke1990principles, lake2017building}. 
Mimicking human intelligence, perceiving environment with composable abstractions have shown consistent improvement in tasks related to systematic generalization~\citep{kuo2020compositional_acl, bogin2021latent_acl, rahaman2021neural_interpreter}, and visual reasoning tasks~\citep{d2021nmn_neurips, assouel2022visual_abstract_reasoning} compared to distributed counterparts.

% ~\citep{greff2020binding}
% In light of those properties, perceiving the scene with composable units is considered as a fundamental aspect of intelligent agent learning. 

Inheriting this spirit, object-centric learning~\citep{burgess19_monet, greff2019iodine, engelcke2019genesis, locatello20_slot_attention} aims to discover a composable abstraction purely from data without external supervision.
% Instead of depicting a scene with a distributed representation, object-centric learning breaks down the scene into a set of latent representations, where each latent is expected to capture a distinct object. 
% Since object concepts should be discovered from data, most of the baselines adopt an auto-encoding framework for the learning process. 
% Under auto-encoding framework, the model is trained to encode the scene into a set of representations and decode them back to the original image. 
Instead of depicting a scene with a distributed representation, it decomposes the scene into a set of latent representations, where each latent is expected to capture a distinct object.
To discover such representation in an unsupervised manner, most existing works employed an auto-encoding framework, where the model is trained to encode the scene into a set of representations and decode them back to the original image. 

% However, minimizing the auto-encoding objective alone cannot guarantee learning composable representations. 
% Auto-encoding objective primarily emphasizes reconstructing the input data, so existing approaches incorporates specially designed inductive bias (\textit{e.g.,} architectural bias~\citep{locatello20_slot_attention} or algorithmic bias~\citep{burgess19_monet, lin2020space, jiang2019scalor}) to guide the object-level disentanglement.

However, the auto-encoding objective is inherently insufficient to learn compositional representation, since maximizing the reconstruction quality does not necessarily requires the object-level disentanglement.
% However, auto-encoding objective is inherently misaligned with the goal of 
To reduce this gap, the existing works incorporate strong inductive biases to further regularize the encoder, such as architectural bias~\citep{locatello20_slot_attention} or algorithmic bias~\citep{burgess19_monet, lin2020space, jiang2019scalor}. 
However, it has been widely observed that these methods are highly sensitive to the choice of hyper-parameters, such as encoder and decoder architectures, and a number of slots, often resulting in suboptimal decompositions by position or partial attributes~\citep{singh21_slate, sajjadi22_osrt, jiang23_lsd} instead of objects. 
Finding the optimal model configuration is also not straightforward in practice due to the missing object labels.
% As a result, 
% Those inductive bias are often sensitive to hyper-parameters, which leads to common failure cases, \textit{e.g.,} dividing the scene into arbitrary shapes rather than objects~\citep{singh21_slate, sajjadi22_osrt, jiang23_lsd}, frequently reported in the field. 
% Considering that object-centric learning aims unsupervised setting, finding a proper hyper-parameters are infeasible in practice. 
% Suspecting that these failure cases stems from a misalignment between the learning objective and the goal of object-centric learning, we argue that learning objective should be aligned with the objective of object-centric representation, \textit{i.e.,} the learning of composable representation.

In this work, we present a novel objective that directly optimizes the compositionality of representations. 
Based upon the auto-encoding framework, our method extracts object representations independently from two distinct images and simulates their composition by the random mixture. 
The composite representation is rendered to an image by the decoder, whose likelihood is evaluated by the generative prior.
The encoder is then jointly optimized to minimize the reconstruction error of the individual images to encode relevant information of the scene (\textit{auto-encoding path}) while maximizing the likelihood of the composite image to ensure the compositionality of the representation (\textit{composition path}). 
Overall, our method can be viewed as extending the conventional auto-encoding approach with an additional regularization on compositionality.
We show that directly injecting compositionality this way significantly boosts the overall quality of object-centric representations and robustness in training.

Our contributions are as follows. 
\textbf{(1)} We introduce a novel objective that explicitly encourages compositionality of representations. 
% To this end, we study strategies of mixing images to simulate composite images and propose a learning objective for maximizing the likelihood of the resulting composite images.
To this end, we investigate strategies to simulate the compositional construction of an image and propose a learning objective for maximizing the likelihood of the composite images.
\textbf{(2)} We evaluate our framework on four datasets and verify that our model consistently surpasses auto-encoding based baselines by a substantial margin. 
\textbf{(3)} We show that our objective enhances the robustness of object-centric learning on three major factors, such as number of latents, encoder and decoder architectures. 

\section{Preliminary}  
\label{sec:preliminary}
\paragraph{Problem setup} 
% \textcolor{blue}{(Introduce general formulation of object-centric learning using auto-encoding)}
\iffalse
Object-centric learning aims to represent an input image $\mathbf{x} \in \mathbb{R}^{H \times W \times C}$ with a set of slot representations $\mathbf{S} \in \mathbb{R}^{N \times D}$, where each slot $\mathbf{s}_n \in \mathbb{R}^{D}$ is expected to capture object-centric concepts. 
Since object concepts should emerge from the data without supervision, a straightforward way is to use an auto-encoding framework to formulate the learning process. 
Formally, the object-centric encoder $E_\theta: \mathbb{R}^{H \times W \times C} \rightarrow \mathbb{R}^{N \times D}$ is trained jointly with a slot decoder $D^{rgb}_\phi:\mathbb{R}^{N\times D} \xrightarrow{}  \mathbb{R}^{H \times W \times C}$ by minimizing the reconstruction loss $d(\mathbf{x},D^{rgb}_\phi(E_\theta(\mathbf{x}))$, where $d$ is a distance metric and typically MSE loss is used. 
\fi
% Object-centric learning aims to represent an input image with a set of composable representations $\mathbf{S} \in \mathbb{R}^{N \times D}$, where each slot $\mathbf{s}_n \in \mathbb{R}^{D}$ is expected to capture object-centric concepts. 
Object-centric learning aims to discover a set of composable representations from an unlabeled image.
Formally, given an image $\mathbf{x} \in \mathbb{R}^{H \times W \times C}$ represented by either RGB pixels or feature from the pre-trained encoder, the objective is to extract the set $\mathbf{S} = \{\mathbf{s}_1,\dots,\mathbf{s}_N\}=E_\theta(\mathbf{x})$, where each element $\mathbf{s}_i\in\mathbb{R}^{D}$ corresponds to the representation of a composable concept (\textit{e.g.}, an object).
Since object concepts should emerge from the data without supervision, a typical approach is to use an auto-encoding framework to formulate the learning process. 
Formally, the object-centric encoder $E_\theta: \mathbb{R}^{H \times W \times C} \rightarrow \mathbb{R}^{N \times D}$ is trained jointly with a decoder $D_\phi:\mathbb{R}^{N\times D} \xrightarrow{}  \mathbb{R}^{H \times W \times C}$ by minimizing the reconstruction loss. 
\begin{equation}
% \theta^*, \phi^* = \argmin_{\theta,\phi} \mathbb{E}_{\mathbf{x}}\left[d(\mathbf{x},D_\phi(E_\theta(\mathbf{x}))\right], 
\mathcal{L}_{\text{AE}}(\theta, \phi)=\mathbb{E}_{\mathbf{x}}\left[d(\mathbf{x},D_\phi(E_\theta(\mathbf{x}))\right], 
\label{eqn:ae_loss}
\end{equation}
where $d$ is a distance metric (\textit{e.g.,} MSE). 
% Below, we introduce the choice of the encoder, decoder, and distance metric used in the literature. 
% Formally, the object-centric encoder $E_\theta: \mathbb{R}^{H \times W \times C} \rightarrow \mathbb{R}^{N \times D}$ is trained jointly with a decoder $D_\phi:\mathbb{R}^{N\times D} \xrightarrow{}  \mathbb{R}^{H \times W \times C}$ by minimizing the reconstruction loss $d(\mathbf{x},D_\phi(E_\theta(\mathbf{x}))$, where $d$ is a distance metric (\textit{e.g.,} MSE). 

\paragraph{Slot Attention Encoder $E_\theta$} 
\label{subsec:slot attention encoder}
\iffalse
Among several instantiations of such framework, Slot Attention Encoder (\cite{locatello20_slot_attention}) provides a promising direction to encapsulate object-centric inductive bias into the encoder. Slot Attention module iteratively refines randomly sampled slot representation through an attention mechanism where each slot competes for the input features. 
\fi
Since the auto-encoding objective is insufficient to learn highly structured representation, the existing approaches incorporate a strong architectural bias in the encoder $E_\theta$ to guide the object-level disentanglement in $\mathbf{S}$.
% Since the auto-encoding objective is insufficient to guide object-level disentanglement in $\mathbf{S}$, the existing approaches incorporate a strong architectural bias in the encoder $E_\theta$.
Among many variants, we consider Slot Attention encoder~\cite{locatello20_slot_attention} due to its popularity and generality. 
\iffalse
A core idea of Slot Attention encoder lies in a that design each slot competes to explain the input features via attention mechanism. Specifically, starting from randomly initialized slots $\mathbf{S}_{init} \in \mathbb{R}^{N\times D}$, slots are iteratively refined by dot-product attention normalized in slot-axis:
\fi
% Slot Attention encoder is designed to 
% to make each latent $\mathbf{s}_i$, i,e., \textit{slot}, competes to explain the input features $\mathbf{z}=f_\theta(\mathbf{x}) \in \mathbb{R}^{M \times D'}$ via attention mechanism, where normalization is applied 
It employs a dot-product attention mechanism between a query (slot) and a key (input), where normalization is applied over the slots by: 
\begin{align}
\label{eqn:attn mask}
\mathbf{A}(\mathbf{x}, \mathbf{S}) = \underset{N}{\text{softmax}} \left(\frac{k(\mathbf{z})\cdot q(\mathbf{S})^T}{\sqrt{D}}\right) \in \mathbb{R}^{M \times N},
\end{align}
where $\mathbf{z}=f_\theta(\mathbf{x}) \in \mathbb{R}^{M \times D'}$ is a flattened input feature encoded by CNN encoder $f_\theta$, and $k,q$ represents linear projection matrices. 
Note that softmax operation is normalized in the query (slots) direction, inducing competition among slots. 
% Employing this attention mechanism, randomly initialized slots $\mathbf{S}^{(0)}\in\mathbb{R}^{N\times D}$ are iteratively refined as:
Based on Equation~\ref{eqn:attn mask}, the slots are iteratively refined by:
\begin{align}
\label{eqn:iterative refine of slot}
% \mathbf{S}^{(n)}=\mathbf{S}^{(n-1)}+ \text{GRU}(\mathbf{W}^T \cdot v(\mathbf{z})), 
% \text{where } \mathbf{W}_{i,j} = \frac{\mathbf{A}_{i,j}}{\sum^M_{j=1}\mathbf{A}_{i,j}}, \mathbf{S}^{(0)} \sim \mathcal{N}(\mu, \text{diag}({\sigma})) \\
\mathbf{S}^{(n+1)}=\text{GRU}(\mathbf{S}^{(n)}, \text{Normalize}(\mathbf{A}(\mathbf{x}, \mathbf{S}^{(n)})^T \cdot v(\mathbf{z}))), 
~~\text{ } \mathbf{S}^{(0)} \sim \mathcal{N}(\mu, \text{diag}({\sigma})).
\end{align}
Here, $\mathbf{S}^{(n)}$ denotes the slot representation after $n$ iterations, $\mu,\sigma$ are learnable parameters characterizing the distribution of the initial slots, $v$ is a linear projection matrix, and $\text{Normalize}(\cdot)$ is a weighted mean operation introduced by \cite{locatello20_slot_attention} to improve stability of the attention.

% $\mathbf{W}$ is a weighted mean of $\mathbf{A}$, which is introduced by \cite{locatello20_slot_attention} to improve stability of attention mechansim.

\paragraph{Slot Decoder $D_\phi$}
\label{subsec:slot decoder}
While the architectural choice for $D_\phi$ is not constrained to a specific form in principle, subsequent works~\citep{singh21_slate, jiang23_lsd} have empirically found that the choice of the decoder crucially impacts the quality of the object-centric representation. 
% Initially \cite{locatello20_slot_attention} proposed a pixel-mixture decoder for $D_\phi$, where each slot is decoded independently, and the resulting images are aggregated in pixel space using alpha-blending. 
\cite{locatello20_slot_attention} proposed a pixel-mixture decoder that renders each slot independently into pixels and combines them with alpha-blending. 
Although slot-wise decoding provides a strong incentive for the encoder to capture distinct objects in each slot, its limited expressiveness hinders its application to complex scenes. 
% To address this issue, \cite{singh21_slate} employed Transformer decoder that takes the entire slots as an input and auto-regressively produces an image using complex interactions among the slots.
To address this issue, \cite{singh21_slate} employed Transformer decoder that takes the entire slots $\mathbf{S}$ as an input and produces an image in an autoregressive manner.
By modeling the complex interactions among the slots, it has shown great improvements in slot representation learning even in complex scenes.
% To address this issue, \cite{singh21_slate} showcases that a transformer decoder significantly improves scalability of object-centric learning. 
% While the architecture of SLATE follows a conventional transformer decoder, it takes the slots $\mathbf{S}$ as inputs through a cross-attention layer. This transformer decoder is trained to reconstruct the image in an autoregressive fashion, which involves minimizing $\min \sum_{i=l}^{l} \text{CrossEntropy}(D_\phi(\mathbf{x}_{<i}, \mathbf{S}), \mathbf{x}_i)$.

Recently, \cite{jiang23_lsd} employed a diffusion model for the slot decoder. 
% In contrast to previous work, \citep{jiang23_lsd} proposes to train an object-centric encoder $E_\theta$ with a denoising objective. 
% In this framework, the object-centric encoder is $E_\theta$ trained with a denosing objective~\citep{}.
Instead of directly reconstructing an input image $\mathbf{x}$, it optimizes the auto-encoding of Equation~\ref{eqn:ae_loss} via denoising objective~\citep{ho20_ddpm} by:
% Specifically, $\mathbf{x}$ is corrupted to a noised image $\mathbf{x}_t$ following the diffusion process and diffusion decoder is trained to correctly predict the original $\mathbf{z}$ as: 
\begin{equation}
\begin{gathered}
\label{eqn:diffusion loss}
\mathcal{L}_{\text{Diff}}(\theta,\phi) = \mathbb{E}_{\epsilon \sim \mathcal{N}(\mathbf{0}, \mathbf{I}), t \sim U(0,1)}\left[w(t)\cdot\|D_\phi(\mathbf x_t,t,\mathbf{S}=E_\theta(\mathbf{x}))-\epsilon\|^2\right], 
% \text{where } \mathbf{x}_t=\sqrt{\bar \alpha_t}\mathbf{x}+\sqrt{1-\bar \alpha_t}\epsilon, t \sim U(0,1), \epsilon \sim \mathcal{N}(\mathbf{0}, \mathbf{I}), \bar \alpha_t=\prod^t_i (1-\beta_i)
\end{gathered}
\end{equation}
where $\mathbf{x}_t=\sqrt{\bar \alpha_t}\mathbf{x}+\sqrt{1-\bar \alpha_t}$ is an corrupted image of an input $\mathbf{x}$ by the forward diffusion process at step $t$, $\bar \alpha_t=\prod^t_i (1-\beta_i)$ is a schedule function, and $w(t)$ is the weighting parameter. In practice, the diffusion decoder is implemented based on UNet architecture~\citep{rombach22_ldm}, where each layer consists of a CNN-layer followed by a slot-conditioned Transformer. Once trained, the decoder generates an image $\mathbf{x}\sim p_\phi(\mathbf{x}|\mathbf{S})$ using iterative denoising, starting from the random Gaussian noise~\citep{ho20_ddpm, rombach22_ldm}. 
Employing a diffusion decoder significantly enhances object-centric representation and generation quality compared to previous arts especially in complex scenes~\cite{jiang23_lsd}. 
% Furthermore, the powerful generative capabilities of the diffusion model substantially improve visual generation quality~\citep{wu2023slotdiffusion}.
% \shcmt{Explain the decoder architecture briefly.}
% \shcmt{Explain how it generates an image briefly.}
% \shcmt{Explain when the diffusion model is better than the others.}

% Note that the loss is reparametrized to predict the noise $\epsilon$ instead of clean $\mathbf{x}$ following \cite{ho20_ddpm}. 
% Although the denoising objective is different from traditional reconstruction loss, we can interpret as a sort of auto-encoding in a perspective that it encodes whole image into few latent representations and recover the whole image from those representations.  

% \subsection{Limitation of auto-encoding objective}
\subsection{Limitations}
\label{subsec:limiation of ae obj}
% \paragraph{Limitation of auto-encoding objective} 
% \textcolor{blue} {(Rely solely on autoencoding does not guarantee good object-centric representation)} \textcolor{blue}{(Discuss what kind of inductive biases in slot attention encourages object representations, even though it is just an auto encoding.)}
While the slot attention with auto-encoding objectives has shown promise in object-centric learning, its success highly depends on the model architectures, such as number of slots and architectures of the encoder and decoder, where suboptimal configuration often leads to dividing the scenes into tessellations~\citep{singh21_slate, sajjadi22_osrt} and objects into the parts~\citep{jiang23_lsd}.
%is often sensitive to the hyperparameters, \textit{e.g.}, architectural choice of encoder and decoder design, and complexity of the dataset, ~\citep{singh21_slate, wu2023slotdiffusion, jiang23_lsd}. 
% It often divides the scenes into tessellations~\citep{singh21_slate, sajjadi22_osrt} and even state-of-the-art object-centric models~\citep{jiang23_lsd} fragments the scene into meaningless subparts in relatively simple scenes. 
However, the optimal model configuration varies depending on the datasets, and discovering them through cross-validation is practically infeasible due to the missing object labels in an unsupervised setting.
We argue that such instability is primarily because the auto-encoding objective is inherently misaligned with the one for object-centric learning, since the former guides the encoder only to minimize the information loss on the input, while the latter demands the object-level disentanglement in the representation, potentially sacrificing the reconstruction quality. 
% The primary objective of the auto-encoding task is to ensure the latent representations capturing relevant information of the given scene. 
% Given a bottleneck of representing a complex scene with a limited number of slots, the model naturally learns to compress highly-corrleated components into each slot representation, thereby encouraging the emergence of the object concepts.
This motivates us to seek an alternative approach that directly encourages object-level disentanglement in the objective function instead of designing architectural biases. 

% When there is a bottleneck on model capacity, the inductive bias in Slot Attention encourages slots to aggregate closely related components into separate slots for efficient auto-encoding. 
% It indicates that the relative capacity of the decoder compared to the data's complexity serves as a strong driving force, promoting the emergence of object concept into distinct slots. 
% However, if this trade-off is not carefully considered, it fails to learn faithful object-centric representation, \textit{e.g}, dividing the scenes into tessellations (\cite{sajjadi22_osrt}). Even state-of-the-art object-centric models \citep{jiang23_lsd} have reported that in relatively simple scenes, the model tends to fragment the scene into meaningless subparts. 

% \section{Method} 
\section{Learning to Compose}
\label{sec:method}

\iffalse
Our goal is to formulate a learning objective that maximizes the compositionality of the slot representation. 
Our main intuition is that arbitrary combination of composable representation are likely to yield another \textit{valid} representation. 
To realize this intuition, our framework is designed to generate composite images by mixing slot representations from two images and maximize its \textit{validity}. 
In this context, we define \textit{validity} as the likelihood of the instance, providing a measure of how realistic the composite image is. 

Overall framework is shown in Figure \ref{fig:overview }. Built upon the conventional auto-encoding path, we design an additional compositional path. In the auto-encoding path, object-centric encoder $E_\theta$ and a slot decoder $D_\phi$ are trained with auto-encoding objective (Equation \ref{eqn:ae_loss}). In compositional path, our primary goal is to generate composite images and maximize the likelihood of these resulting composite images. In Section~\ref{subsec:slot mixing}, we discuss how we mix two sets of slot representations extracted from two distinct images and generate composite image. Then, in Section~\ref{subsec:compsitional objective}, we illustrate how we maximize the likelihood of the composite image. As we do not have a ground-truth image for the composite image, we will exploit a critic function that can provide a learning signal to maximize the likelihood of the composite image. 
\fi

Our goal is to improve object-centric learning by modifying its objective function to be more directly aligned with learning compositional slot representation than the auto-encoding loss.
% Our intuition is that arbitrary combination of composable repres entation are likely to yield another \textit{valid} representation. 
Our main intuition is that arbitrary compositions of object representation are likely to yield another \textit{valid} representation. 
To realize this intuition, our framework is designed to generate composite images by mixing slot representations from two images and maximize their validity measured by the data prior. 

\begin{figure}[!t]
\centering
\includegraphics[width=1.0\textwidth]{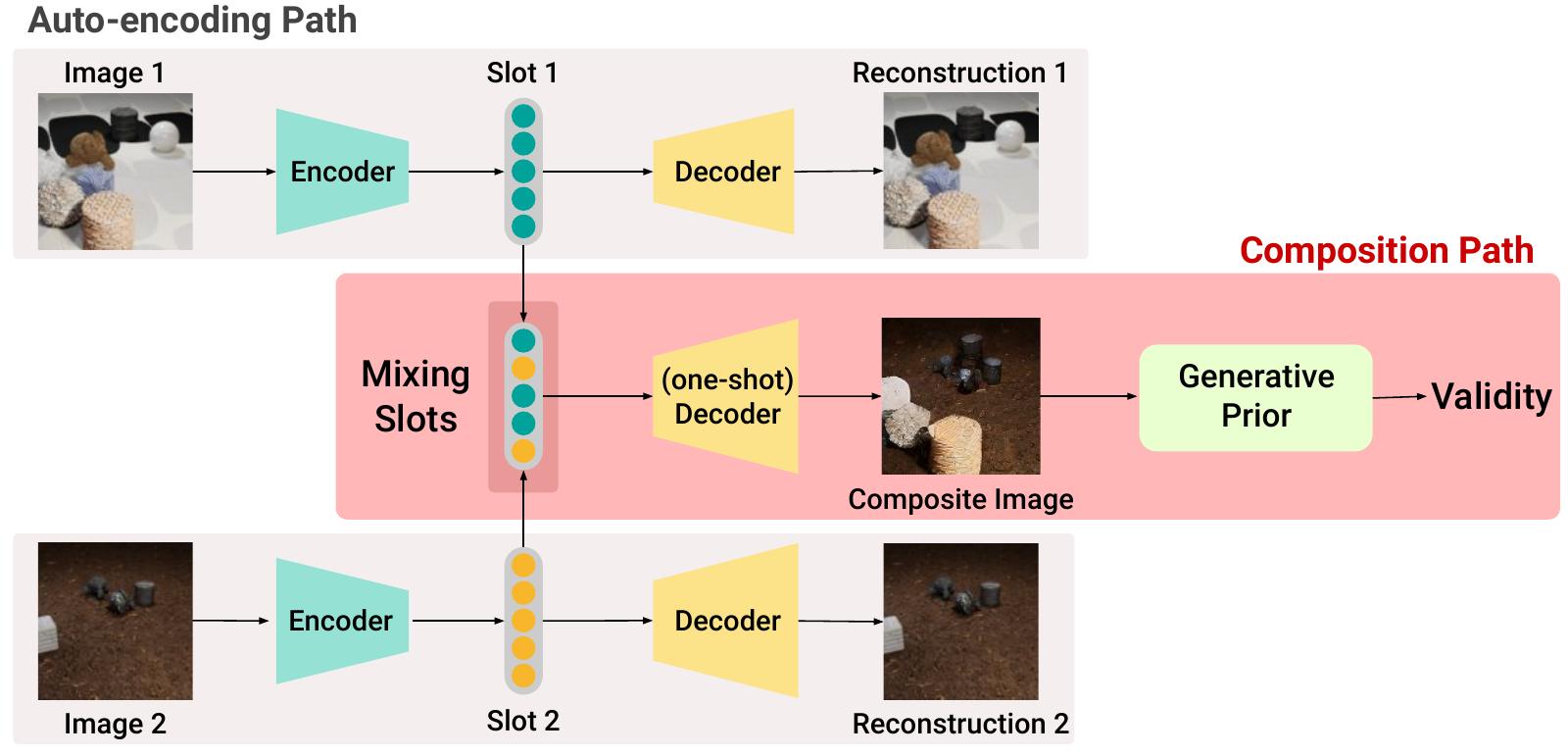}
\caption{
\textbf{Overview of our method.}
Our framework consists of two paths: an auto-encoding path and a composition path. 
The auto-encoding path ensures slot representations encode relevant information about an image. In contrast, the composition path encourages the compositionality of the representations by constructing the composite representation through the mixture of slots from two separate images (Section~\ref{subsec:slot_mixing}), and assessing the quality of the composite image by the generative prior (Section~\ref{subsec:compsitional_objective}). 
The encoder is jointly optimized by both paths.
\iffalse
Our framework consists of two path; an auto-encoding path (conventional object-centric learning approach) and a composition path. 
While the auto-encoding path ensures slot representations encoding relevant information of given images, 
the composition path assesses the compositionality of the slot representations. 
Specifically, we first encode two distinct images $\mathbf{x}^1$ and $\mathbf{x}^2$ into $\mathbf{S}^1,\mathbf{S}^2$, respectively, and mix those slots to generate a composite image $\mathbf{x}^c$ (Section~\ref{subsec:slot_mixing}). Then, we employ a generative prior to maximize the validity of $\mathbf{x}^c$ (Section~\ref{subsec:compsitional_objective}).
\fi
% \shcmt{Highlight more on example images (current level of details is fine but less highlight the models). This figure is to deliver conceptual idea of our framework and we will use another figure to describe the detailed components, such as one-shot decoder, diffusion decoder, gradient flow, etc.}
% 
}
\label{fig:overview}
\vspace{-0.5cm}
\end{figure}

Figure~ \ref{fig:overview} illustrates the overall framework of our method.
Our framework is built upon the conventional object-centric learning that learns both the slot encoder and decoder by the auto-encoding path on individual images (Section~\ref{sec:preliminary}).
To impose compositionality on slot representation, we incorporate an additional \textit{composition path} that constructs a composite slot representation from two images by the mixing strategy (Section~\ref{subsec:slot_mixing}) and assesses the quality of the image generated from the mixed slots by the generative prior (Section~\ref{subsec:compsitional_objective}).
% {\color{red}The gradient signals from the composition path are used to update only the encoder $E_\theta$ to prevent suspicious collaboration between the encoder and decoder in generating composite images from suboptimal slots.}\shcmt{we may move it to other section. It is sufficient to say that the encoder is trained to optimize the composition path.}
This way, the auto-encoding path ensures that each slot contains the relevant information of an input image, while such slots are constrained to capture composable components of the scenes (\textit{e.g.,} objects) by regularizing the encoder through the composition path.
%object-level disentanglement in such slots are enforced by the composition path. 
% The slot encoder is then jointly trained with auto-encoding path to enhance 

% Built upon the conventional auto-encoding path, we design an additional compositional path. In the auto-encoding path, object-centric encoder $E_\theta$ and a slot decoder $D_\phi$ are trained with auto-encoding objective (Equation \ref{eqn:ae_loss}). In compositional path, our primary goal is to generate composite images and maximize the likelihood of these resulting composite images. In Section~\ref{subsec:slot mixing}, we discuss how we mix two sets of slot representations extracted from two distinct images and generate composite image. Then, in  Section~\ref{subsec:compsitional objective}, we illustrate how we maximize the likelihood of the composite image. 

\subsection{Mixing Strategy for composing slot representation}
\label{subsec:slot_mixing}
% This section illustrates how we design to compose two slot representations and generate composite image $\mathbf{x}^c$. 
Given $\mathbf{S}^1, \mathbf{S}^2 \in \mathbb{R}^{N \times D}$ extracted from two distinct images $\mathbf{x}^1, \mathbf{x}^2$, we construct their composite slot representation $\mathbf{S}^{c}\in \mathbb{R}^{N \times D}$ by
\begin{align} 
\label{eqn:slot composition}
\mathbf{S}^{c} = \pi(\mathbf{S}^1, \mathbf{S}^2), 
% \mathbf{x}^c = D_\phi(\mathbf{S}^c).
\end{align}
where $\pi(\cdot, \cdot)$ denotes a composition function of two sets. 
The primary role of the composition function is to simulate potential combinations of slot-wise compositions. 
Since our goal is to maximize the compositionality of unseen slot combinations, the composition function should be capable of exploring a broad range of compositional possibilities. 
Below, we introduce simple instantiations of such function. 
% While $\pi(\cdot, \cdot)$ is not limited to a specific form, we introduce a possible instanstiation of this function. 

\paragraph{Random Sampling}
In this approach, we randomly sample $N$ slots among $2N$ slots \textit{i.e.}, $\mathbf{S}^{c} \overset{N}{\sim} (\mathbf{S}^1 \cup \mathbf{S}^2)$. 
As it explores over all of the possible combinations, this composition function encourages the slot representation itself to be highly composable to generate valid images for any combinations.
% In terms of augmentation, it is also beneficial to generate maximally diverse samples for the composite image, boosting reguarlization on composite path.
On the other hand, it may produce invalid combinations of slots on rare occasions, \textit{e.g.}, omitting the background slots or sampling two objects placed in the same location. 
% While this kind of aggressive search promotes compositionality, it comes with the risk of encountering inherently invalid combination of slots, e.g., omitting the slot containing the background information or sampling two objects located on the same location. 

\paragraph{Sharing Slot initialization}
\label{sec:noise_share}
% One way to mitigate such suspicious compositions is to avoid sampling similar slots from two images at the same time. 
One way to mitigate such suspicious compositions is to constrain $\mathbf{S}^c$ to be valid composition of the scene.
However, strictly ensuring this constraint is non-trivial due to the stochastic nature of slot attention \textit{i.e.}, each slot is sampled stochastically from its underlying distribution and the association between the slots and scenes varies depending on the initialization.
Instead, we adopt a rather simple approach that employs the identical slot initialization $\mathbf{S}^{(0)}$ in Equation~\ref{eqn:iterative refine of slot} for two images, and sample the exclusive set of slots.
Formally, let $I_1$ and $I_2$ be a random partition of slot indices  \textit{i.e.}, $I_1\cup I_2=\{1,...,N\}, I_1\cap I_2=\emptyset$.
Then we construct the composite slot by $\mathbf{S}^c=\mathbf{S}^1_{I_1}\cup \mathbf{S}^2_{I_2}$, where $\mathbf{S}^1$ and $\mathbf{S}^2$ are slots extracted by Equation~\ref{eqn:iterative refine of slot} from $\mathbf{x}^1$ and $\mathbf{x}^2$, respectively, which are initialized with the same $\mathbf{S}^{(0)}$.%while sharing the same initialization $\mathbf{S}^{(0)}$.
The underlying intuition is that the slot initialization is reasonably correlated with the objects it captures (Figure~\ref{fig:slot_init_share_appendix}), hence sampling from exclusive slots is likely to be valid scenes than the random sampling.

\subsection{Maximizing likelihood of the composite image}
\label{subsec:compsitional_objective}
\iffalse
In this section, we present how we maximize the likelihood of $\mathbf{x}^c$. Since there is no ground-truth image available for $\mathbf{x}^c$, we employ a generative model to serve as a critic function, that can update $\mathbf{x}^c$ in a direction that maximizes the $p(\mathbf{x}^c)$. 
Among the various candidate models, we opt for a diffusion model as our critic function. The choice of a diffusion model is motivated by recent study demonstrating its notable generative capacity at scale, along with its substantial mode coverage~\citep{xiao21_denoising_diffusion_gans, huang21_variational_diffusion, song21_maximum_likelihood_score_based_diffusion}. While variational Autoencoder (VAEs) can offer direct likelihood measurement capabilities, they frequently struggle to model complex scenes effectively. \textcolor{blue}{(TODO : find a drawback of autoregressive transformer..)}. Consequently, we decide to employ a diffusion model for the critic function across all datasets.
\fi
Given the composite slot $\mathbf{S}^c$ obtained by the previous section, our next step is quantifying its validity \textit{i.e.}, measuring how valid the composition of two image slots is.
To this end, we decode it back to an image by $\mathbf{x}^c=D_\phi(\mathbf{S}^s)$ and measure the likelihood of the image using the generative prior $p(\mathbf{x}^c)$.

\paragraph{Generative Prior}
\iffalse
While the choice of a generative model learning $p(\mathbf{x}^c)$ is flexible, we opt for a diffusion model. 
The choice of a diffusion model is motivated by its remarkable generative performance across various datasets~\citep{ho20_ddpm, rombach22_ldm, ramesh2022dalle-2} and its substantial mode coverage~\citep{xiao21_denoising_diffusion_gans}.
While we can train an external diffusion model for a generative prior, we employ a diffusion decoder $D_\phi$. 
This decoder can be trained in the auto-encoding path and also serves as a score estimator in the composition path, where "score" refers to the gradient of the log-likelihood. Specifically, a diffusion decoder $D_\phi$ is trained in auto-encoding path by minimizing the denoising objective in Equation \ref{eqn:diffusion loss}. 
\fi
To model the generative prior $p(\mathbf{x}^c)$, we opt for a diffusion model~\citep{ho20_ddpm} due to its excellence in generation quality and mode coverage~\citep{xiao21_denoising_diffusion_gans}.
The latter is especially important in our framework since the model evaluates the prior over potentially out-of-distribution samples generated by the composition (Section~\ref{subsec:slot_mixing}).
% Instead of introducing a pre-trained diffusion model for a generative prior, we reuse the diffusion decoder $D_\phi$ employed in the auto-encoding path (Section~\ref{sec:preliminary}), which greatly improves the parameter-efficiency and memory. 
Instead of introducing an additional pre-trained diffusion model, we employ the diffusion-based decoder in the auto-encoding path (Section~\ref{sec:preliminary}), and reuse it as a generative prior.
This way, our decoder $D_\phi$ is trained by minimizing the reconstruction loss by denoising objective in Equation~\ref{eqn:diffusion loss}, while serving as a generative prior in the composition path. 
It greatly improves the parameter-efficiency and memory, and the need for pre-trained generative prior per dataset. 

\paragraph{Maximizing $p(\mathbf{x}^c)$}
Given the generative prior, we maximize the likelihood $p(\mathbf{x}^c)$ with respect to $\mathbf{x}^c$ in the composition path. 
Since $\mathcal{L}_{\text{Diff}}$ in Equation \ref{eqn:diffusion loss} is minimizing the upper bound of negative log likelihood of $x^c$~\citep{ho20_ddpm}, minimizing $\mathcal{L}_{\text{Diff}}$ with respect to $\mathbf{x}^c$ leads to the maximization of the likelihood $p(\mathbf{x}^c)$. 
% Since $\mathcal{L}_{\text{Diff}}$ in Equation \ref{eqn:diffusion loss} is equivalent to the upper bound of negative log likelihood of $x^c$~\citep{ho20_ddpm}, minimizing $\mathcal{L}_{\text{Diff}}$ with respect to $\mathbf{x}^c$ leads to the maximization of the likelihood $p(\mathbf{x}^c)$.
However, computing the gradient of $\mathcal{L}_{\text{Diff}}$ requires expensive computation of Jacobian maxtrix of the decoder and it often degrades the overall training stability.
Following \citep{poole22_dreamfusion}, the gradient of $\mathcal{L}_{\text{Diff}}$ with respect to $\theta$ can be approximated as: 
% \begin{align}
% \label{eqn:sds loss}
% \nabla_{\theta} \mathcal{L}_{\text{Prior}}(\theta) = \frac{\partial \mathbf{x}^c}{\partial \theta} \nabla_{\mathbf{x}^c} \mathcal{L}_{\text{Diff}}(\theta,\phi) \approx \mathbb{E}_{t, \epsilon}[w(t)(D_\phi(\mathbf{x}^{c}_t,t,\mathbf{S}^{c})-\epsilon)\frac{\partial \mathbf{x}^c}{\partial \theta}]. 
% \end{align}
\begin{align}
\label{eqn:sds loss}
\nabla_{\theta} \mathcal{L}_{\text{Prior}}(\theta) = \mathbb{E}_{t, \epsilon}[w(t)(D_\phi(\mathbf{x}^{c}_t,t,\mathbf{S}^{c})-\epsilon)\frac{\partial \mathbf{x}^c}{\partial \theta}]. 
\end{align}
where $\epsilon\sim\mathcal{N}(\mathbf{0},\mathbf{I})$ is a noise, $t\sim\mathcal{U}(t_{\text{min}}, t_{\text{max}})$ is a timestep, respectively, $w(t)$ is a weighting function dependent to $t$, and $\mathbf{x}_t^c=\sqrt{\bar \alpha_t}\mathbf{x}^c+\sigma_t\epsilon$ is a corrupted image of $\mathbf{x}^c$ from forward diffusion process. 
By updating the encoder parameters $\theta$ with $\nabla_{\theta} \mathcal{L}_{\text{Prior}}$, $\mathbf{x}^c$ is guided toward high probability density region following the diffusion prior. 
% Note that the gradient signals from Equation~\ref{eqn:sds loss} are used to update only the encoder $E_\theta$ to prevent suspicious collaboration between the encoder and decoders in generating composite images from suboptimal slots. 
Note that optimization of the Equation~\ref{eqn:sds loss} is with only respect to the encoder parameter while fixing the decoder.
It prevents suspicious collaboration between the encoder and decoders in generating composite images from suboptimal slots. 

% \paragraph{Constructing $\mathbf{x}^c$ with deterministic decoder $D_\psi$}
\paragraph{Surrogate One-Shot Decoder}
As discussed earlier, our framework exploits the diffusion model $D_\phi$ as a decoder and generative prior in the auto-encoding and composition paths, respectively.
% One drawback is that generating the composite image $\mathbf{x}^c$ by the diffusion requires iterative denoising process, which takes significant time and makes the backpropation through the decoder non-trivial.
One drawback is that the diffusion decoder requires an iterative denoising process to generate the composite image $\mathbf{x}^c$, which takes significant time and makes the backpropagation through the decoder non-trivial.
To address this problem, we employ a one-shot decoder $D_\psi$ as a surrogate for $D_\phi$ to support fast and differentiable decoding of $\mathbf{x}^c$.
\footnote{We also consider one-step denoising result of the diffusion decoder using Tweedie's formula~\citep{stein1981tweedie_estimation, robbins1992tweedie_empirical} but observe severe degradation in performance due to its inferior quality.}

We employ a bidirectional Transformer~\citep{devlin2018bert} that takes the composite slot $\mathbf{S}^c$ and the learnable mask tokens $\mathbf{m}\in\mathbb{R}^{HW\times C}$ as input, and produces the composite image by a single forward process by $\mathbf{x}^c=D_\psi (\mathbf{m}, \mathbf{S}^c)$.
The decoder is trained along with the auto-encoding path by:
\begin{align}
\label{eqn:recon loss}
\mathcal{L}_{\text{Recon}}(\theta, \psi) = ||D_\psi(\mathbf{m}, E_\theta(\mathbf{x})) - \mathbf{x}||^2.
\end{align}
Note that the generation quality of the one-shot decoder $D_\psi$ is behind the powerful diffusion decoder $D_\phi$, and serves only to compute the $\mathbf{x}^c$ in Equation~\ref{eqn:sds loss}.
We observe that such weak decoder is sufficient to compute the meaningful gradient through the Equation~\ref{eqn:sds loss}, presumably because the gradients are accumulated over various noise levels $t$.

\subsection{Learning Objective}
\label{subsec:overall_objective}
 
In this section, we summarize the overall framework and objective function.
Our framework consists of two paths; auto-encoding path and composition path. 
In auto-encoding path, encoder $E_\theta$ and two different decoders $D_\phi, D_\psi$ are trained to minimize auto-encoding objective in Equation~\ref{eqn:diffusion loss} and Equation~\ref{eqn:recon loss}. 
% Note that those two decoders have clearly different roles. 
% The diffusion decoder $D_\phi$ serves as a main decoder, which is responsible for generating final outputs. 
% As described in Section~\ref{subsec:slot decoder}, final images are sampled using $D_\phi$ through an iteratively denoising process, starting from random Gaussian Noise. 
% On the other hand, $D_\psi$ acts as a surrogate for $D_\phi$, 
% with the primary goal of facilitating one-step decoding from composite slots $\mathbf{S}^c$ to an image $\mathbf{x}^c$ in composition path. 
In composition path, we first extract $\mathbf{S}^c$ with Equation~\ref{eqn:slot composition} and generate $\mathbf{x}^c$ with the deterministic decoder $D_\psi$, and update the encoder to maximize the Equation~\ref{eqn:sds loss} while fixing decoders $D_\phi$ and $D_\psi$. 
% Then, $D_\psi$ maximizes the likelihood $p(\mathbf{x}^c)$ with Equation~\ref{eqn:sds loss}. 
% In addition to the objectives outlined earlier, we empirically find that incorporating a regularization term on the slot attention mask enhances object-centric representations:
We find that incorporating an additional regularization term on the slot attention mask is helpful in enhancing object-centric representations:
\begin{equation}
\label{eqn:regularization loss}
\mathcal{L}_{\text{Reg}}(\theta) = \mathbf{A}^1 \cdot \text{sg}(||\mathbf{x}^1 - \mathbf{x}^c||^2) + \mathbf{A}^2\cdot \text{sg}(||\mathbf{x}^2 - \mathbf{x}^c||^2), 
\end{equation}
% where $\mathbf{A}^1=\mathbf{A}(\mathbf{x}^1, \mathbf{S}^{1^{(n)}}), \mathbf{A}^2=\mathbf{A}(\mathbf{x}^1, \mathbf{S}^{2^{(n)}})$ are attention masks 
where $\mathbf{A}^1=\mathbf{A}(\mathbf{x}^2, \mathbf{S}^{1^{(n)}}), \mathbf{A}^2=\mathbf{A}(\mathbf{x}^1, \mathbf{S}^{2^{(n)}})$ are attention masks from the last iteration of slot attention for $\mathbf{x}^1, \mathbf{x}^2$  (Equation~\ref{eqn:attn mask}), respectively, and $\text{sg}(\cdot)$ denotes stop-gradient operator. 
% This regularization term encourages $E_\theta$ to attend more on the object region.
It encourages the source and the composite images to be consistent over the object area captured by the slots, enhancing the content-preserving composition.
% thereby facilitating the learning of object-centric representations. 
The overall objective is then formulated as follow: 
\begin{equation}
\begin{gathered}
\label{eqn:total loss}
\mathcal{L}_{\text{Total}}(\theta,\phi,\psi) = \lambda_{\text{Prior}}\mathcal{L}_{\text{Prior}}(\theta)+ \lambda_{\text{Diff}}\mathcal{L}_{\text{Diff}}(\theta,\phi)+ \lambda_{\text{Recon}}\mathcal{L}_{\text{Recon}}(\theta,\psi) + \lambda_{\text{Reg}}\mathcal{L}_{\text{Reg}}(\theta)\end{gathered}
\end{equation}
% \begin{align}
% \label{eqn:total loss}
% \mathcal{L}_{total} = \lambda_{critic}\mathcal{L}_{critic} + \lambda_{lsd} \mathcal{L}_{\text{Diff}} + \lambda_{\text{Recon}} \mathcal{L}_{\text{Recon}}+ \lambda_{Reg} \mathcal{L}_{Reg},
% \end{align}
where $\lambda_{\text{Prior}}, \lambda_{\text{Diff}}, \lambda_{\text{Recon}}, \lambda_{\text{Reg}}$ are hyperparameters for controlling the importance of each term. We empirically find that $\lambda_{\text{Prior}}=\lambda_{\text{Diff}}=\lambda_{\text{Recon}}=1.0, \lambda_{\text{Reg}}=0.25$ generally works well and use it throughout the experiments. 
% \begin{equation}
% \begin{gathered}
% \label{eqn:total loss2}
% \theta^* = \argmin_{\theta} \lambda_{critic}\mathcal{L}_{critic}+ \lambda_{lsd}\mathcal{L}_{\text{Diff}}+ \lambda_{\text{Recon}}\mathcal{L}_{\text{Recon}} + \lambda_{Reg}\mathcal{L}_{Reg}, \\
% \phi^* = \argmin_{\phi} \lambda_{lsd}\mathcal{L}_{\text{Diff}}, \psi^* = \argmin_{\psi} \lambda_{\text{Recon}}\mathcal{L}_{\text{Recon}}, 
% \end{gathered}
% \end{equation}

\section{Related Work}

\paragraph{Object-centric learning}
% Object-centric learning aims to represent the scene with multiple latent representations capturing a distinct object entities purely from data. Due to the unsupervised nature of the problem, auto-encoding framework is widely-chosen for formulating the learning process~\citep{burgess19_monet, greff2019iodine, engelcke2019genesis, engelcke2021genesisv2, lin2020space, jiang2019scalor, eslami2016air, crawford2019spair}. 
The most dominant paradigm of object-centric learning is employing the auto-encoding objective~\citep{burgess19_monet, greff2019iodine, engelcke2019genesis, engelcke2021genesisv2, lin2020space, jiang2019scalor, eslami2016air, crawford2019spair}. 
To guide the model to learn structured representation under reconstruction loss, \cite{locatello20_slot_attention} introduces Slot Attention, where each slot is iteratively refined with dot-product attention mechanism normalized in slot direction, inducing competition between the slots. 
Follow-up studies~\citep{singh21_slate, seitzer22_dinosaur, sajjadi22_osrt} demonstrate that Slot Attention with an auto-encoding objective has the potential to attain object-wise disentanglement even in complex scenes. 
Nonetheless, auto-encoding alone often involves training instability, which leads to attention-leaking problem~\citep{kim2023slash}, or dividing the scene into Voronoi tessellations~\citep{sajjadi22_osrt, jiang23_lsd}. 
% We suspect that these inherent limitations stem from lack of direct optimization for learning object-centric representation. 
To overcome such challenges, there have been a few attempts on revising the learning objective such as replacing image reconstruction loss with denoising objective~\citep{jiang23_lsd, wu2023slotdiffusion} or contrastive loss~\citep{henaff2022odin, wen2022slotcontrast}. 
Nevertheless, these approaches still do not impose direct learning of object-centric representations.

\cutparagraphup
\paragraph{Generative Prior}
There are increasing interests in exploiting the knowledge pre-trained from generative prior to various applications such as solving inverse problems~\citep{chung2023solving3dinverse}, guidance in conditional generation~\citep{graikos2022diffusion_plug_play, liu2023semantic_diffusion_guidance}, and image manipulations~\citep{ruiz2023dreambooth, zhang2023inversion_based_style_trasnfer, ruiz2023hyperdreambooth}. 
One prominent approach in this direction is text-to-3D Generation, where a large-scale pre-trained 2D diffusion model~\citep{rombach22_ldm, saharia2022imagen} is leveraged to generate realistic 3D data without ground-truth~\citep{wang22_sjc, lin2023magic3d, metzer2023latentnerf, wang2023vsd}. 
The seminal work by \citep{poole22_dreamfusion} formulates a loss based on a probability density distillation to distill a pre-trained 2D image prior to a 3D model. 
Back-propagating the loss through a randomly initialized 3D model, \textit{e.g.}, NeRF~\citep{mildenhall2021nerf},
the model gradually updates to generate high-fidelity 3D renderings. 
Inspired by this line of work, we employ a generative model in our approach to maximize the validity of the given images. 
% \vspace{1cm}

 \section{Experiment}
% We evaluate the strength of our novel compositional objective across four object-centric datasets. 
% To conduct a comprehensive analysis, we examine an unsupervised object segmentation task~\citep{locatello20_slot_attention, jiang23_lsd} and verify robustness of our objective to various hyper-parameters including encoder architecture, decoder capacity and number of slots. 
% We then conduct an ablation study to identify the specific contribution of each component in our framework. 
% As our results will demonstrate, our method consistently outperforms state-of-the-art performance across all datasets and under varying parameter configurations.
\paragraph{Implementation Details}
We base our implementation on existing frameworks~\citep{singh21_slate,jiang23_lsd}.
We employ the features from the pre-trained auto-encoder\footnote{https://huggingface.co/stabilityai/sd-vae-ft-ema-original} to represent an image.
For the slot encoder, we employ the CNN based on UNet architecture~\citep{singh22_steve,jiang23_lsd} to produce a high-resolution attention map.
Also, we employ an implicit Slot Attention~\citep{chang2022implicit_sa} to stabilize the iterative refinement process in slot attention. 
% For a bi-directional transformer decoder $D_\psi$, we adapt the slot-conditioned transformer architecture described in \cite{singh21_slate}. 
% A key difference is that it does not utilize causal masking, enabling one-step decoding. 
For the slot mixing strategy, we opt for a sampling with sharing slot initializations for all the experiments unless specified, since it shows slightly better performance than the random sampling strategy.
When we compute $\mathcal{L}_{\text{Prior}}$ (Equation~\ref{eqn:sds loss}), we use $t_{\text{min}}=0.02, t_{\text{max}}=0.5$ following a recent report in \citep{wang2023vsd} that employing too high noise level impairs the optimization. 
% Finally, the number of slots for each dataset are set to sum of maximum number of ground-truth objects and one (background). 

\cutparagraphup
\paragraph{Datasets}
We validate our method on four datasets. %, each presenting different levels of difficulty.
% CLEVRTex < PTR ~= MSN < Super-CLEVR
% complex on shapes.  Super-CLEVR -> material, reflection
CLEVRTex~\citep{karazija2021CLEVRTex} consists of various rigid objects with homogeneous textures. 
MultiShapeNet~\citep{stelzner2021multishapenet} includes more complex and realistic furniture objects. 
PTR~\citep{hong2021ptr} and Super-CLEVR~\citep{li2023superclevr} contain objects composed of multi-colored parts and textures. 
All of the datasets are center-cropped and resized to 128x128 resolution images. 
 % \jhcmt{give some evidence that implies LSD may fail on PTR, CLEVRTex, MSN.}
% \textcolor{blue}{(Todo : Explain Super-CLEVR and justify why autmentation is applied. )}
. 

\cutparagraphup
\paragraph{Baselines}
\iffalse
We compare our method against two strong baselines in the literature, 
SLATE~\citep{singh21_slate} and LSD~\citep{jiang23_lsd}, which employ the autoregressive Transformer and diffusion model as a decoder, respectively. 
% For fair comparison, we employ the same encoder architecture based on slot attention~\citep{locatello20_slot_attention} in all compared methods including ours.
For fair comparison, all baselines share the same encoder architecture with our method.
For LSD and our method, we employ the same pre-trained auto-encoder~\footnote{https://huggingface.co/stabilityai/sd-vae-ft-ema-original}~\citep{rombach22_ldm} to represent an input image.
Since SLATE runs on discrete features, we employ the pre-trained VQGAN model~\citep{esser2021vqgan}, and denotes it as SLATE+.
\fi
We compare our method against two strong baselines in the literature, 
% SLATE~\citep{singh21_slate} that employing the autoregressive transformer as a decoder, and LSD~\citep{jiang23_lsd} based on diffusion decoder.
SLATE~\citep{singh21_slate} and LSD~\citep{jiang23_lsd}, which employ autoregressive Transformer and diffusion-based decoders, respectively. 
% Note that our method without composition path can be reduced to LSD.
% Note that LSD can be considered as a special case of our method without composition path.
Note that our method without composition path reduces to LSD.
For a fair comparison, we employ the same encoder architecture based on slot attention~\citep{locatello20_slot_attention} in all compared methods including ours.
For LSD and our method, we employ the same pre-trained auto-encoder~\citep{rombach22_ldm} to represent an input image.
Since SLATE runs on discrete features, we employ the features from the pre-trained VQGAN model~\citep{esser2021vqgan} and denote it as SLATE+.
All baselines including ours are trained for 200K iterations.
% Both methods share the same encoder based on slot attention~\citep{locatello20_slot_attention} while employing different decoders such as Transformer and diffusion, respectively.
% Both models are representative work in auto-encoding based slot attention variants. 
% LSD, a state-of-the-art model in object-centric learning, is a baseline for direct comparison with our approach~\footnote{Due to lack of officially supported implementation, we reproduced LSD faithfully following the paper.}.
% The only difference between LSD and our model is existence of composition path, allowing us to directly assess the impact of our compositional objective.
% Note that  only auto-encoding path in our framework is equivalent to LSD model. 
% For computational efficiency, both LSD and our model employed large-scale pretained auto-encoders~\footnote{https://huggingface.co/stabilityai/sd-vae-ft-ema-original}~\citep{rombach22_ldm} and the diffusion model is trained on the latent space. 
% SLATE+ is another competitive baseline, which use an autoregressive transformer decoder. SLATE+ is a variant of SLATE~\citep{singh21_slate}, where original vae model of SLATE is replaced by large-scale pretrained VQGAN model~\citep{esser2021vqgan} for a fair comparison with our method. 

% As the performance of SLATE+ consistently outperforms SLATE, we only include SLATE+ for our baseline.

\cutparagraphup
\paragraph{Evaluation Metrics}
\iffalse
Following previous works~\citep{jiang23_lsd, singh21_slate, singh22_steve, chang2022implicit_sa}, we evaluate the models with Adjusted Rand Index for foreground objects (FG-ARI), mean Intersection over Union (mIoU) and mean Best Overlap (mBO). 
While FG-ARI measures a cluster similarity of foreground objects, mIoU and mBO assess segmentation qualities by measuring overlap between a ground-truth mask and a predicted mask. 
When measuring mIoU, we use Hungarian Algorithm~\citep{kuhn1955hungarian} to find best matching between ground-truth object masks and predicted masks.  
In contrast, we assign a ground-truth object mask to each slot having the largest overlap when evaluating mBO. 
To evaluate those metrics, we extract attention masks (Equation~\ref{eqn:attn mask}) from the last iteration of the slot attention and take argmax operation along the slot dimension to get hard segmentation masks. 
\fi
Following the previous works~\citep{jiang23_lsd, singh21_slate, singh22_steve, chang2022implicit_sa}, we report the unsupervised segmentation performance with three measures: Adjusted rand index for foreground objects (FG-ARI), mean intersection over union (mIoU), and mean best overlap (mBO). 
These metrics measure the overlap between the slot attention masks and ground-truth object masks,
where FG-ARI focuses more on the coverage of the object area.
% each of which is measured based on the slot attention mask and ground-truth object masks. 

% !TEX root = main.tex
% without JSD
\begin{figure*}[!t]
    \begin{minipage}{\textwidth}
    \captionsetup{type=table}

\centering
\begin{footnotesize}
    \caption{\textbf{Comparison results on unsupervised object segmentation.}
    We evaluate the how well the slot attention masks coincide with the ground-truth objects using FG-ARI, mIoU, and mBO (The higher is better).
    All results are evaluated on held-out validation set.
    % We measure the accuracy of the predicted slot attention mask on the held-out validation images.
    % In all measures, the higher is better.
    % Our method consistently outperforms the baselines in all datasets and measures.
    % We assess the segmentation quality across various datasets with FG-ARI, mIoU, and mBO. Our model outperforms all baselines on all datasets in terms of FG-ARI, mIoU, and mBO.
    }
    \vspace{-0.5em}
    % CLEVRTex
    \centering
    \begin{subtable}{0.49\textwidth}
        \subcaption{CLEVRTex}
        % \centering
        \vspace{-1mm}
        \begin{tabular}{lccc}
        \toprule
        Model   & FG-ARI    & mIoU  & mBO   \\
        \midrule
        SLATE+  & 71.29    & 52.04  & 52.17 \\
        LSD     & 76.44    & 72.32  & 72.44 \\
        Ours    & \textbf{93.06}    & \textbf{74.82}  & \textbf{75.36} \\
        \bottomrule
        \end{tabular}
    \end{subtable}
    % CLEVRTex
    \centering
    \begin{subtable}{0.49\textwidth}
        \subcaption{MultiShapeNet}
        % \centering
        \vspace{-1mm}
        \begin{tabular}{lccc}
        \toprule
        Model   & FG-ARI    & mIoU  & mBO   \\
        \midrule
        SLATE+  & 70.44    & 15.55  & 15.64 \\
        LSD     & 67.72    & 15.39  & 15.46 \\
        Ours    & \textbf{89.8}    & \textbf{59.21}  & \textbf{59.4} \\
        \bottomrule
        \end{tabular}
    \end{subtable}\\
    % PTR
    \vspace{0.7em}
    % \begin{subtable}{0.49\textwidth}
    \centering
    \begin{subtable}{0.49\textwidth}
        \subcaption{PTR}
        % \centering
        \vspace{-1mm}
        \begin{tabular}{lccc}
        \toprule
        Model   & FG-ARI    & mIoU  & mBO   \\
        \midrule
        SLATE+  & \textbf{91.25}    & 14.1	& 14.22 \\
        LSD     & 61.1    & 10.18  & 10.33 \\
        Ours    & 90.65  & \textbf{40.89}  & \textbf{41.45} \\
        \bottomrule
        \end{tabular}
    \end{subtable}
    \begin{subtable}{0.49\textwidth}
        \subcaption{Super-CLEVR}
        \centering
        \vspace{-1mm}
        \begin{tabular}{lccc}
        \toprule
        Model   & FG-ARI    & mIoU  & mBO   \\
        \midrule
        SLATE+  & 43.73 &	29.12 &	29.49 \\
        LSD     & 54.79 &	14.12 &	14.43 \\
        Ours    & \textbf{63.08} &	\textbf{47.17} &	\textbf{48.03} \\
        \bottomrule
        \end{tabular}
    \end{subtable}
    \label{tab:main} 
    % \vspace{-2mm}
    \vspace{0.25cm}
\end{footnotesize}
\end{minipage}

   \begin{minipage}{\textwidth}
    \centering
    \includegraphics[width=1.0\textwidth]{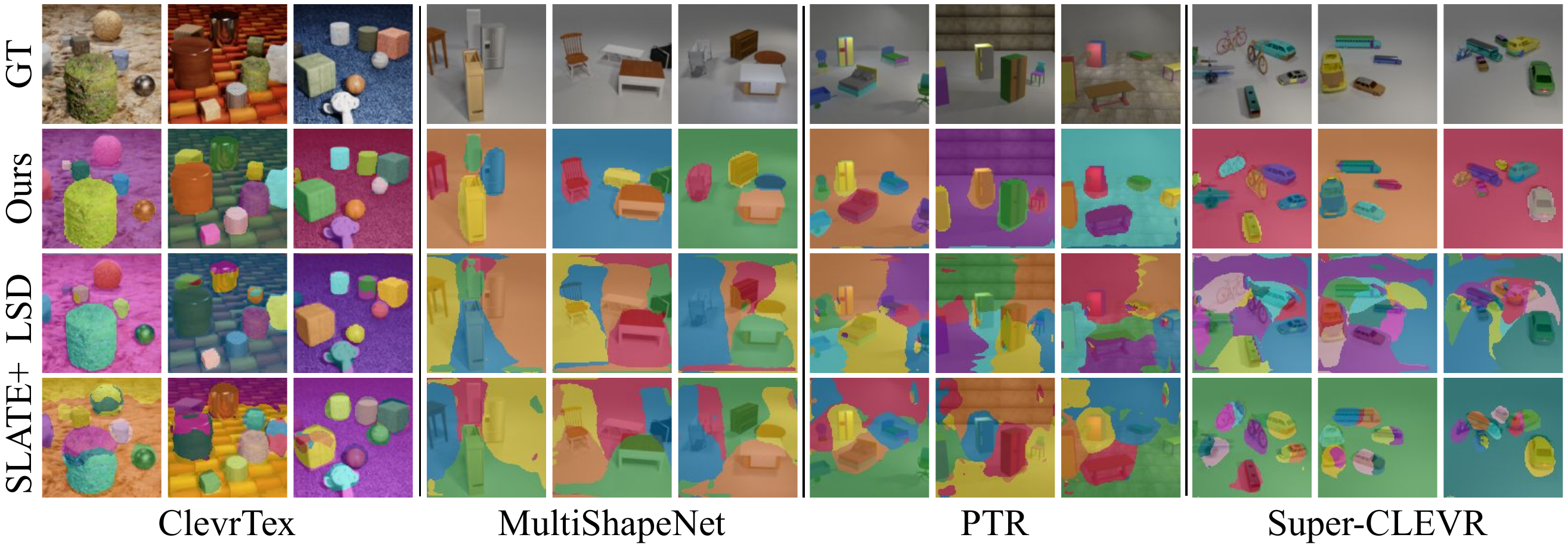}% change it to the figure
        \vspace{-0.5cm}
    % \vspace{2cm}
    % put figure here
    % \vspace{2cm}
    \captionsetup{type=figure}
    \caption{\textbf{Qualitative results on unsupervised object segmentation.} 
    The baselines tend to split an object into different slots (CleverTex) and/or combine different objects and background into a single (MultiShapeNet, PGR, Super-CLEVR). 
    On the other hand, our method produces consistently better object masks, showing improved disentanglement of objects and background in all datasets. 
    More results are presented in the Figure~\ref{fig:unsupervised_seg_appendix}.
    \textbf{Zoom in} for better view.}
    % \shcmt{increase the font size in the figure}}  
    \label{fig:figure2_qual}    
    \end{minipage}
    \vspace{-1cm}
\end{figure*}

\begin{figure}

\end{figure}
\subsection{Unsupervised Object Segmentation}
\iffalse
% In this section, we evaluate our method on an unsupervised object segmentation task. 
Our results in Table~\ref{tab:main} demonstrates that our method significantly outperforms the baselines. 
Our model surpasses the baselines on three out of four datasets in FG-ARI and outperforms all of the baselines by a significant margin in terms of mIoU and mBO. 
For qualitative analysis, we visualize the segmentation masks in Figure~\ref{fig:figure2_qual}. 
SLATE+ frequently split the foreground object masks into multiple segments in CLEVRTex and Super-CLEVR dataset and fails to capture meaningful object concepts in PTR and MultiShapeNet. Similarly, LSD fails to segment the object region across all datasets except CLEVRTex dataset and divides the scene with position bias.  
% Especially in PTR and Super-CLEVR datasets, it fails to capture objects and tends to divide the scene with positions. 
Note that the reconstruction loss and denoising loss of the baselines are decreased enough to reconstruct the original image (see Figure~\ref{}, Figure~\ref{}). 
We argue that this is a direct evidence of misalignment between the auto-encoding objective and the goal of achieving optimal object-centric representation. 
In contrast, our method consistently produces sharp and holistic masks for each object across all datasets.
Considering that the only difference between LSD and our model is existence of compositional objective, the significant gap between LSD and ours verifies the effectiveness of our novel objective. 
\fi
We first present the comparison results of our method with baselines on unsupervised object segmentation.
% Table~\ref{tab:main} and Figure~\ref{fig:figure2_qual} summarize the quantitative and qualitative comparison results, respectively.
Table~\ref{tab:main} summarizes the quantitative results.
% Overall, our method consistently outperforms the baselines in almost all datasets and measures.
Our method significantly improves the FG-ARI scores over the baselines in all datasets (8 to 29\% improvement) except PTR, indicating that it captures an object holistically into an individual slot while the baselines tend to split the object into multiple parts and distribute it across multiple slots. %and encode them into the different representations.
In terms of mIoU and mBO, our method improves the baselines over all datasets, especially when the background is monolithic (MultiShapeNet, PTR, and Super-CLEVR).
It indicates that the baselines struggle to separate the objects from the background when there exists a strong correlation between them, while our method can still robustly identify the objects.
Overall, the results indicate that our method consistently outperforms the baselines by a significant margin. 
Notably, the consistent and significant improvement over LSD indicates that our regularization on the compositionality is effective in learning object-centric representation. 
% Overall, our method consistently outperforms the baselines in all datasets and measures, except the PTR dataset, where the SLATE+ slightly outperforms in terms of FG-ARI.
% Specifically, our method outperforms the baselines in terms of FG-ARI 
% higher FG-ARI scores in our method shows that it tends to capture the objects 

We also present the qualitative results in Figure~\ref{fig:figure2_qual}.
It shows that SLATE frequently splits the foreground object masks into multiple segments in CLEVRTex and Super-CLEVR datasets, and fails to capture object entities in PTR and MultiShapeNet. 
Similarly, LSD fails to segment the object in all datasets except CLEVRTex dataset, and tends to rely on positional bias in PTR and Super-CLEVR.
In contrast, our method consistently captures objects with tight boundaries.
% It shows the similar trends in quantitative evaluation, where the baselines are splitting the object into multiple parts, and often not able to clearly distinguish the object from the background. 
% Notably, it shows that LSD separates objects based on spatial biases in some datasets (PTR and Super-CLEVR), showing that 

% \subsection{Robustness of Our Objective}

\begin{figure}[t!]
  \begin{subfigure}{0.28\textwidth} % Adjust the width as needed
    \centering
    \includegraphics[width=\linewidth]{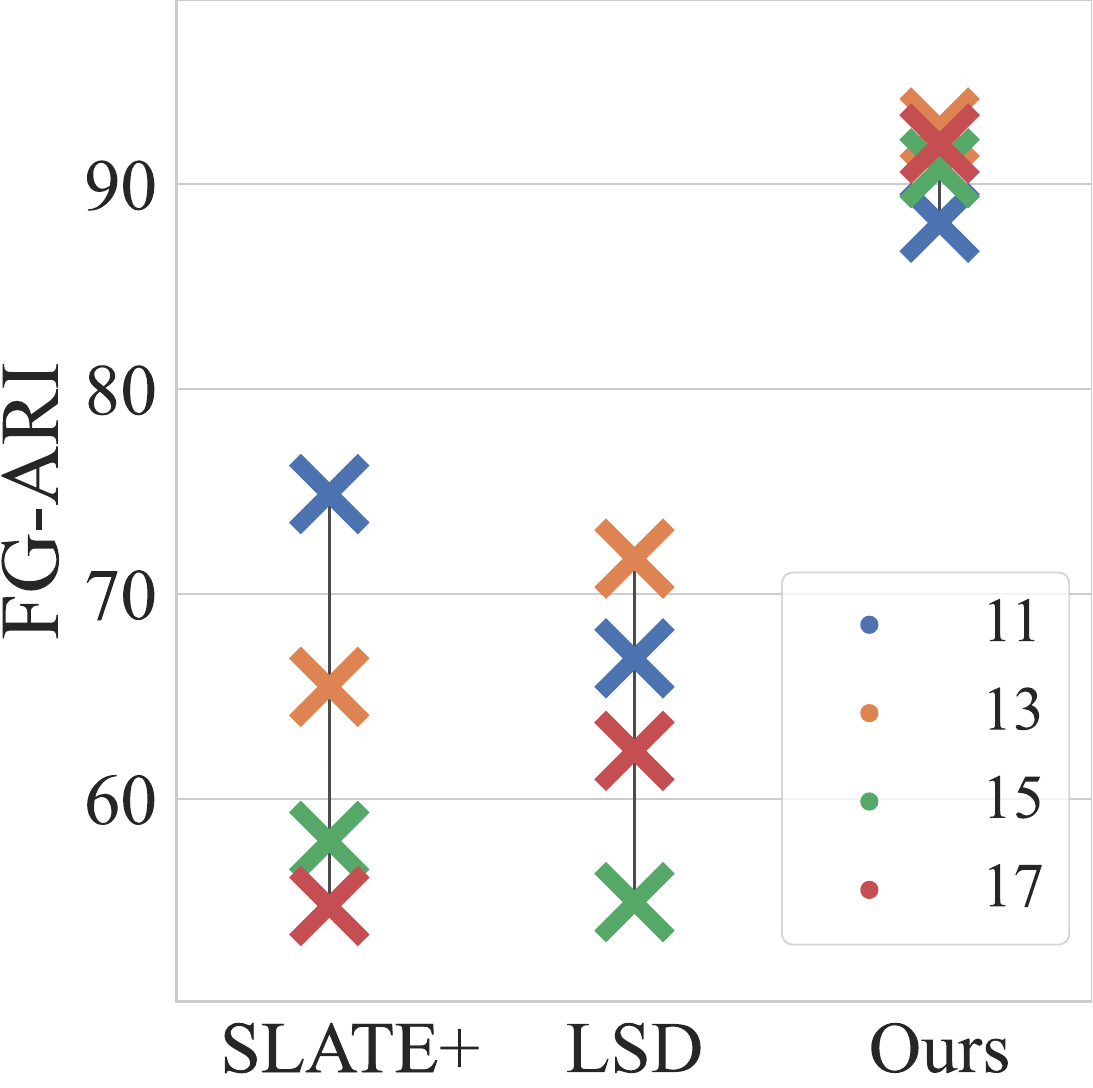}
    \caption{Number of slots}
    \label{subfig:robust_num_slot}
  \end{subfigure}
\hspace{0.2cm}
  \hfill
  \begin{subfigure}{0.28\textwidth} % Adjust the width as needed
    \centering
    \includegraphics[width=\linewidth]{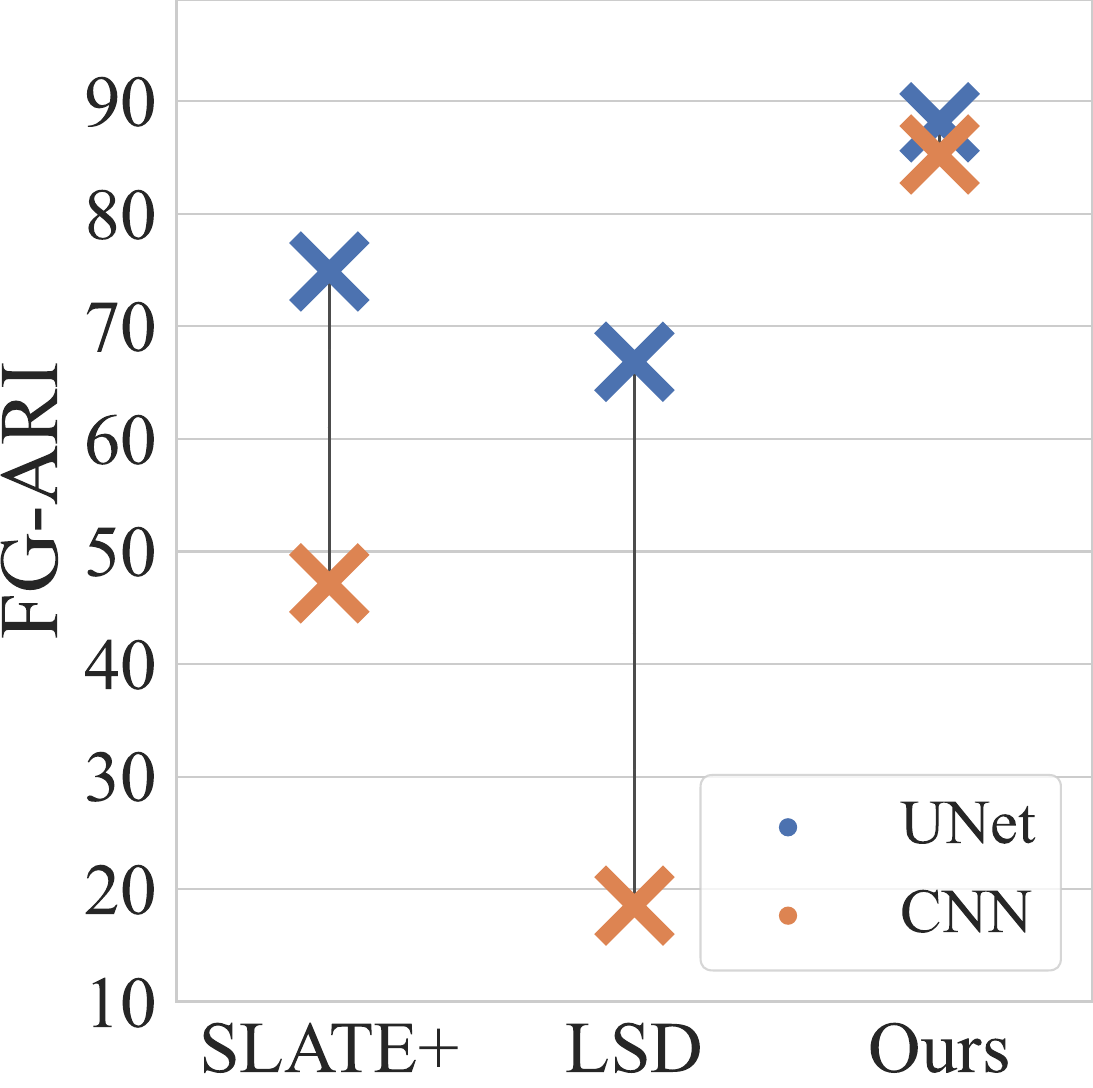}
    \caption{Encoder architecture}
    \label{subfig:robust_enc_type}
  \end{subfigure}
  \hspace{0.2cm}
  \hfill
  \begin{subfigure}{0.28\textwidth} % Adjust the width as needed
    \centering
    \includegraphics[width=\linewidth]{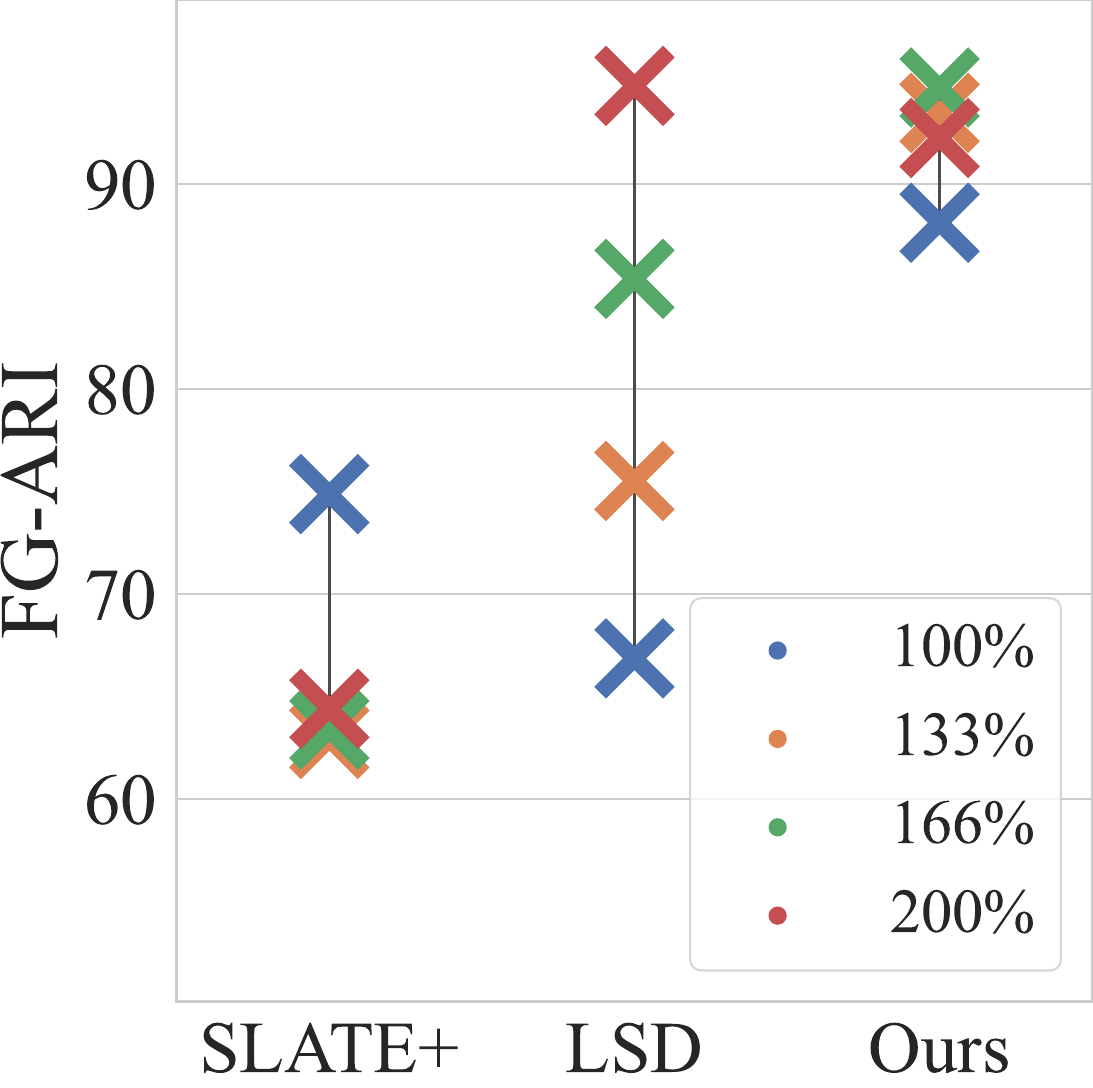}
    \caption{Decoder capacity}
    \label{subfig:robust_dec_cap}
  \end{subfigure}
  \caption{\textbf{Robustness against various architectural biases.} We evaluate the robustness of our model various different number of slots, encoder architectures, and decoder capacities. 
  %Among various hyperparameters, our model steadily shows powerful performance against baselines. 
  % See full results in Figure~\ref{fig:full_ablation_appendix}.
  Results based on mIoU and mBO are presented in Figure~\ref{fig:full_ablation_appendix}.
  }
  \label{fig:robustness_all}
  \vspace{-0.5cm}
\end{figure}

\subsection{Robustness of Compositional Objective}
\label{subsec:robustness_test}
\cutparagraphup
Compared to approaches based on auto-encoding, our method directly incorporates the objective to learn compositional representation, thus is more robust to the choice of architectural biases and hyperparameters. 
To demonstrate this, we evaluate our method while varying three major factors that are known to be highly sensitive in the previous approaches, such as number of slots, encoder architecture, and decoder capacity.
% To this end, we evaluate our method while varying architectural choices and hyperparameters, such as number of slots, encoder and decoder architectures, and compare the results against the baselines.
Figure~\ref{fig:robustness_all} summarizes the result on CLEVRTex dataset based on FG-ARI. %, where the results based on mIoU and mBO are presented in Figure~\ref{fig:full_ablation_appendix} in the Appendix.
All methods are trained up to 100K iterations for fair comparison.
% \shcmt{Training iteration?}

\cutparagraphup
\paragraph{Number of slots}
Since object-centric learning assumes no prior knowledge on data, the mismatch between the number of objects and slots is inevitable in practice. 
To evaluate such robustness, we vary the number of slots from 11 to 17.
Figure~\ref{subfig:robust_num_slot} presents the result.
% It shows that the performance of the baselines are highly sensitive to the number of slots, where SLATE tends to deteriorates more with an increasing number of slots.
It shows that the performance of the baselines is highly sensitive to the number of slots. Specifically, SLATE tends to deteriorate more as the number of slots increases.
Compared to the baseline, our method achieves more robust performance by encoding an object into a slot while leaving excess slots empty. 
% utilizing empty slots when there is more slots than the objects.
% Compared to the baselines, our method shows much robust performance, successfully  
% We explore four different numbers of slots (11, 13, 15, 17).

\cutparagraphup
\paragraph{Encoder architecture}
To identify the effect of slot encoder, we consider two popular architectures in the literature; 
a multi-layer CNN encoder~\citep{singh22_steve} and UNet-based encoder~\citep{Ronneberger15_unet}.
Figure~\ref{subfig:robust_enc_type} summarizes the result.
It shows that employing the weaker encoder generally deteriorates the performance of the baselines significantly, indicating that architectural bias in the encoder is critical in the auto-encoding objective.
Interestingly, the performance of our method is hardly affected by such drastic modifications, showing great robustness.
% It shows that directly regularizing the compositionality in the objective is important.
% It is noteworthy that FG-ARI, mIOU, mBO values of two baselines are substantially decreased with multi-layer CNN encoder, while our model maintains comparable performance.

\cutparagraphup
\paragraph{Decoder capacity}
It is widely observed that the choice of decoder is also crucial in object-centric learning, since the highly expressive decoder can often bypass the object representation to minimize the reconstruction loss~\citep{singh21_slate}.  
To examine such effect, we gradually increase the feature dimensions of the decoder to 133$\%$, 166$\%$, and 200$\%$.
    Figure~\ref{subfig:robust_dec_cap} summarizes the result.
It shows that increasing the decoder capacity hampers the performance in SLATE. 
LSD exhibits the opposite trends showing a large improvement in FG-ARI, although its performance drops significantly in mIoU (Figure~\ref{fig:full_ablation_appendix}). 
Compared to the baselines, our method is much less sensitive to the decoder capacity, while the performance tends to improve slightly with increased capacity in all measures. 

% Overall, the above results suggest that the quality of object-centric representation is highly dependent on various factors in appraoches based on auto-encoding objective. 
% In contrast, our model consistently achieves the oustanding performance with minimal variance across all configurations, even in radical changes in the encoder architecture. 
% It shows that addressing the misalignment in objective of auto-encoding and object-centric learning has crucial impact, and is practically more useful.
Overall, the results indicate that the quality of object-centric representation is significantly influenced by various factors in the auto-encoding-based methods.
Conversely, our model consistently delivers outstanding performance across all configurations, even with major alterations to the encoder architecture.
% This demonstrates that rectifying the misalignment between the objectives of auto-encoding and object-centric learning is crucial and practically beneficial.
It demonstrates that our regularization through the composite path can directly encourage the model to learn compositional representation, greatly enhancing robustness to architectural biases.

\begin{table}[t!]
    \centering
    \caption{\textbf{Ablation study on CLEVRTex dataset.}
    All models are trained up to 100K iterations.
    % We investigate the importance of each component within our framework. Each of the component clearly contributes to the improvement. All models are evaluated at 100k iterations.
    }    
    \vspace{-0.1cm}
    \begin{tabular}{ccc|ccc}
    \toprule
    $\mathcal{L}_\text{Prior}$ &
    $\mathcal{L}_\text{Reg}$ & Share $\mathbf{S}^{(0)}$ & FG-ARI & mIoU & mBO  \\
    \midrule
    % \cmark & & & & & 39.78 & 51.72 & 51.94  \\
    \xmark & \xmark & \xmark & 42.48 & 52.26 & 52.41  \\
    \hline
    \cmark  & \xmark & \xmark & 65.76 & 67.72 & 67.62  \\
    \cmark & \xmark & \cmark & 70.29 & 69.08 & 69.28  \\
    \xmark  & \cmark & \cmark & 65.26 & 58.81 & 58.99  \\
    % \cmark & \cmark &  \xmark  & \textbf{88.8} & 73.36 & 73.72  \\
    \hline
    \cmark & \cmark & \cmark & \textbf{88.15} & \textbf{75.30} & \textbf{75.64}  \\
    \bottomrule
    \end{tabular}
    \label{tab:ablation}
\end{table}

\begin{figure}[!t]
\centering
\includegraphics[width=0.99\linewidth]{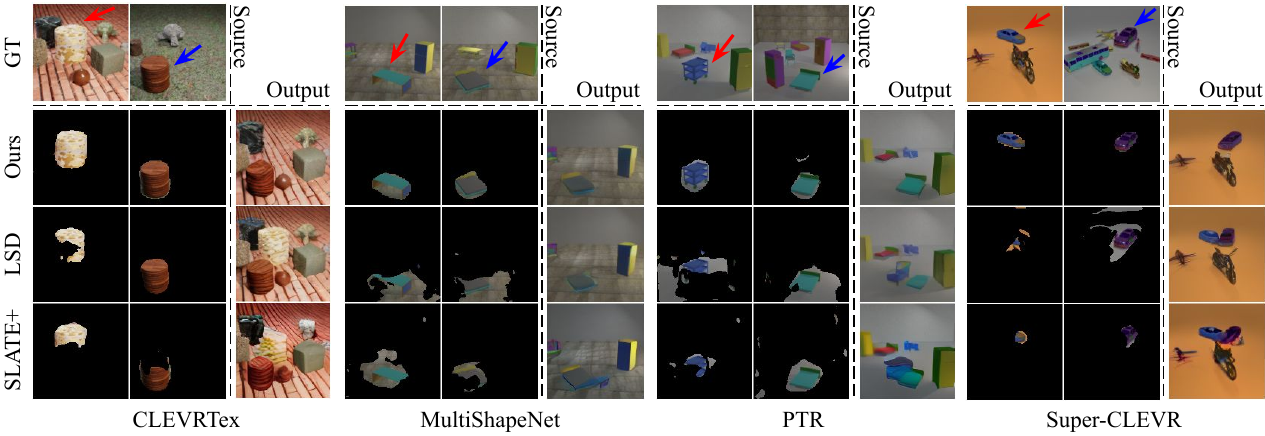}
\vspace{-0.1in}
\caption{\textbf{Investigating object representation through compositional generation.} 
We investigate the compositionality of learned representations by removing ({\color{red}red} arrow) and adding ({\color{blue}blue} arrow) object slots between two images and generating the composite image. 
More results are in Figure~\ref{fig:Composition_two_imgs_appendix}.
% We investigate the compositionality of learned representations with compositional generation. 
% We synthesize the composite image by replacing a {\color{red} slot} from an image with the other {\color{blue} slot} from a different image. 
}
\label{fig:compositional_gen}
\vspace{-0.5cm}
\end{figure}

\subsection{Internal Analysis}
\label{subsec:ablation_study}
\paragraph{Component-wise Contributions}
To identify the contributions of each component in our framework, we conduct an ablation study and present the result in Table~\ref{tab:ablation}. 
% All of the experiments are conducted on CLEVRTex dataset and evaluated at 100k iterations. 
The first row corresponds to our model with only the auto-encoding path, while the last row is the complete version of our model. 
% Comparing the first row with subsequent rows, incorporating the composition path significantly improves overall metrics compared to the auto-encoding counterparts.
Comparing the first row with the others shows that incorporating the composition path significantly improves overall quality.
Adding $\mathcal{L}_{\text{Prior}}$, we observe a substantial improvement in all three metrics. 
Considering that FG-ARI measures the correct cluster membership of pixels within the objects, increased FG-ARI indicates that the generative prior encourages the encoder to capture more holistic object representations. 
This is because the generative prior penalizes the encoder for fragmenting the objects, thereby discouraging the generation of unrealistic partial objects in the composite image. 
Comparing the second and the third rows, we observe that sharing the slot initialization $\mathbf{S}^{(0)}$ slightly enhances mIoU and mBO scores. 
This improvement is likely attributed to the increased training stability by avoiding invalid slot combinations as shown in Figure~\ref{fig:slot_init_share_appendix}. 
% Finally, additionally exploiting $\mathcal{L}_{\text{Reg}}$, our model achieves the state-of-the-art performance. 
% The regularization term encourages the encoder to attend locally on object regions. 
% It is noteworthy that employing $\mathcal{L}_{\text{Reg}}$ alone without generative prior (fourth row) shows significant degradation in performance. 
% This is because the encoder get any penalties even if it learns to copy-and-paste the partial object into composite image. 
Incorporating regularization $\mathcal{L}_{\text{reg}}$ alone in the composition path does not improve the performance (fourth row), while combined with generative prior, it leads to significant improvement.
% In summary, our model achieves state-of-the-art performance by incorporating all of the proposed components. 

\paragraph{Compositional Generation}
% \begin{figure}[t!]
%     \centering
%     \includegraphics[width=1.0\textwidth]{figures/robust_all.pdf}
%     \caption{\textbf{Robustness across Various Hyperparameters.} We evaluate the robustness of our model across different capacities of decoder and number of slots. Among various hyperparameters, our model steadily shows powerful performance against baselines. }
%     \label{fig:robustness}
% \end{figure}

\iffalse
To demonstrate the impact of our compositional objective, we present a compositional generation example in Figure~\ref{fig:compositional_gen}. 
Given two images on the top row in the figure, we encode those images into slots (from second to fourth rows) and mix them to generate the composite image on the rightmost columns. 
Images in the second to fourth rows indicate attention masks from the slot attention encoder, representing which region the object is attending to.  
While compositional generation of our model is not only realistic but also faithfully reflects the original objects in the source images, the composite image in SLATE+ includes a suspicious object, \textit{e.g.,} the object within a red circle. This is because the slot in the third column (red box) contains a fragmented object that leads to an unrealistic object. 
In our model, the generative prior penalizes such impaired composite image, which in turn motivates the encoder to capture a holistic object instead of encoding only partial objects. 
Please see Section~\ref{} for more qualitative results. 
\fi
% To investigate the impact of our method, we compare the composition generation results. 
We present the compositional generation results to further investigate the impact of our composition path.
Figure~\ref{fig:compositional_gen} presents the results. 
Given two images, we construct the composite representation by replacing one object slot from the first image (red arrow) to another from the second image (blue arrow), and producing the image by the decoder.
Based on visualization of the learned slots, we observe that the baselines often fail to learn compositional slot representation, by separating objects into multiple slots or encoding background with an object.
It leads to failures in object-level manipulation, such as retaining an object after the removal (LSD in MultiShapeNet and PTR), altering the content of the added object (SLATE in MultiShapeNet), or transforming background with the object (SLATE in PTR and LSD in Super-CLEVR).
In contrast, our method produces both semantically meaningful and realistic images from composite slot representations, supporting our claim that we can regularize object-centric learning through the proposed compositional path.

% \subsection{Compositional Generation} \textcolor{blue}{}

\section{Conclusion}
\label{sec:conclusion}
\cutsectiondown
In this paper, we introduced a method to address the misalignment between object-centric learning and the auto-encoding objective. 
Our method is based on auto-encoding framework, and incorporates an additional branch to directly assess the compositionality of the representation. 
This involves constructing composite representations from two separate images and optimizing the encoder jointly with the auto-encoding path to maximize the likelihood of the composite image. 
Despite the simplicity, our extensive experiments demonstrate that our framework consistently improves the object-centric learning over the auto-encoding frameworks. 
It also shows that our method greatly enhances the robustness to the choice of architectural biases and hyperparameters, which typically pose sensitivity challenges in auto-encoding-centric approaches. %which are often too sensitive in approaches based on auto-encoding objective.
% Through the extensive experiments on four datasets and various model configurations, we demonstrate that simply 
% To directly guide the model to learn compositional representation, we build our framework based 
% Based on auto-encoding framework, we propose to incorporate an additional constraint that an arbitrary composition of object representations should be valid

\paragraph{Acknowledgements} 
This work was supported in part by Institute of Information \& communications Technology Planning \& Evaluation (IITP) grant (No.2022-0-00926, 2022-0-00959, 2021-0-02068, and 2019-0-00075) and National Research Foundation of Korea(NRF) grant (2021R1C1C1012540 and 2022R1C1C1009443) funded by the Korea government(MSIT). 

% Sungjin Ahn was supported by Young Researcher Program (No. 2022R1C1C1009443) through the National Research Foundation of Korea (NRF) funded by the Ministry of Science and ICT. 


\bibliography{main}
\bibliographystyle{main}

\clearpage
\appendix

% \section{Appendix}
\section{Additional Implementation Details}  
% \subsection{Datasets}
% We use four different datasets. 
% \subsection{Architecture Details}
Table~\ref{tab:hyperparams of model} provides details of hyperparameters used in experiments. 
For the Slot Attention encoder $E_\theta$ and a diffusion decoder $D_\phi$, we base our implementation on \cite{jiang23_lsd}. 
Specifically, in the Slot Attention encoder, we employ a CNN-based UNet image encoder.
Prior to the UNet encoder, we incorporate a single layer CNN to downsample the original $128 \times 128$ image to a $64\times 64$ image. 
Implementing the diffusion decoder $D_\phi$, we follow the design of the LSD decoder. 
The overall structure of $D_\phi$ is based on the U-Net architecture, where each layer is composed of CNN layers and a transformer layer.
The surrogate decoder $D_\psi$ is implemented with the Transformer Architecture in \cite{singh21_slate}. It takes slots as input through cross-attention layers. 
In the experimental setting, we augment the Super-CLEVR dataset by randomly altering the background color to another color.

\begin{table}[h!]
\centering
\footnotesize
\begin{tabular}{@{}llcccc@{}}
\toprule
%                \cmidrule(l){3-6} 
% Module      & Hyperparameter     & CLEVRTex       & MSN    & PTR    & Super-CLEVR                   \\ \midrule
General     &   Batch Size       & 64 \\
            &   Training Steps   & 200K \\
            &   Learning Rate    & 0.0001 \\
    \midrule
CNN Backbone             &  Input Resolution &     128    \\
                        &  Output Resolution &     64    \\
                        &  Self Attention  &     Middle Layer    \\
                        &  Base Channels &     128    \\
                        &  Channel Multipliers &     [1,1,2,4]    \\
                        &  \# Heads &     8    \\
                        &  \# Res Blocks / Layer &     2    \\
                        &  Slot Size  & 192     \\ \midrule
                        % &  Learning Rate  & 0.0001       \\ \midrule
Slot Attention          &  Input Resolution &     64    \\
                        &  \# Iterations  &     7    \\
                        &  Slot Size  &     192 \\ \midrule
                        % &  Learning Rate  &     0.0001       \\ \midrule
Auto-Encoder            &  Model &     KL-8 \\
                        &  Input Resolution &     128 \\
                        &  Output Resolution &     16 \\
                        &  Output Channels &     4 \\ \midrule
Diffusion Decoder          &  Input Resolution &     16 \\
                        &  Input Channels &     4 \\
                        &  $\beta$ scheduler   &     Linear \\
                        &  Mid Layer Attention   &     Yes \\
                        &  \# Res Blocks / Layer &     2    \\
                        % &  Learning Rate   &     0.0001    \\
                        &  \# Heads & 8  \\
                        &  Base Channels & 192 \\
                        &  Attention Resolution &     [1,2,4,4]    \\
                        &  Channel Multipliers & [1,2,4,4] \\ \midrule
Surrogate Decoder    
                        & Layers & 8 \\
                        & \# Heads & 8 \\
                        & Hidden Dim & 384 \\ \midrule

\end{tabular}
\caption{\textbf{Hyperparameters used in our experiments.}}
\label{tab:hyperparams of model}
\end{table}

\section{Additional Results}  
\subsection{Additional Results on Robustness Tests}
\begin{figure}[t!]
  \begin{subfigure}{0.3\textwidth} % Adjust the width as needed
    \centering
    \includegraphics[width=\linewidth]{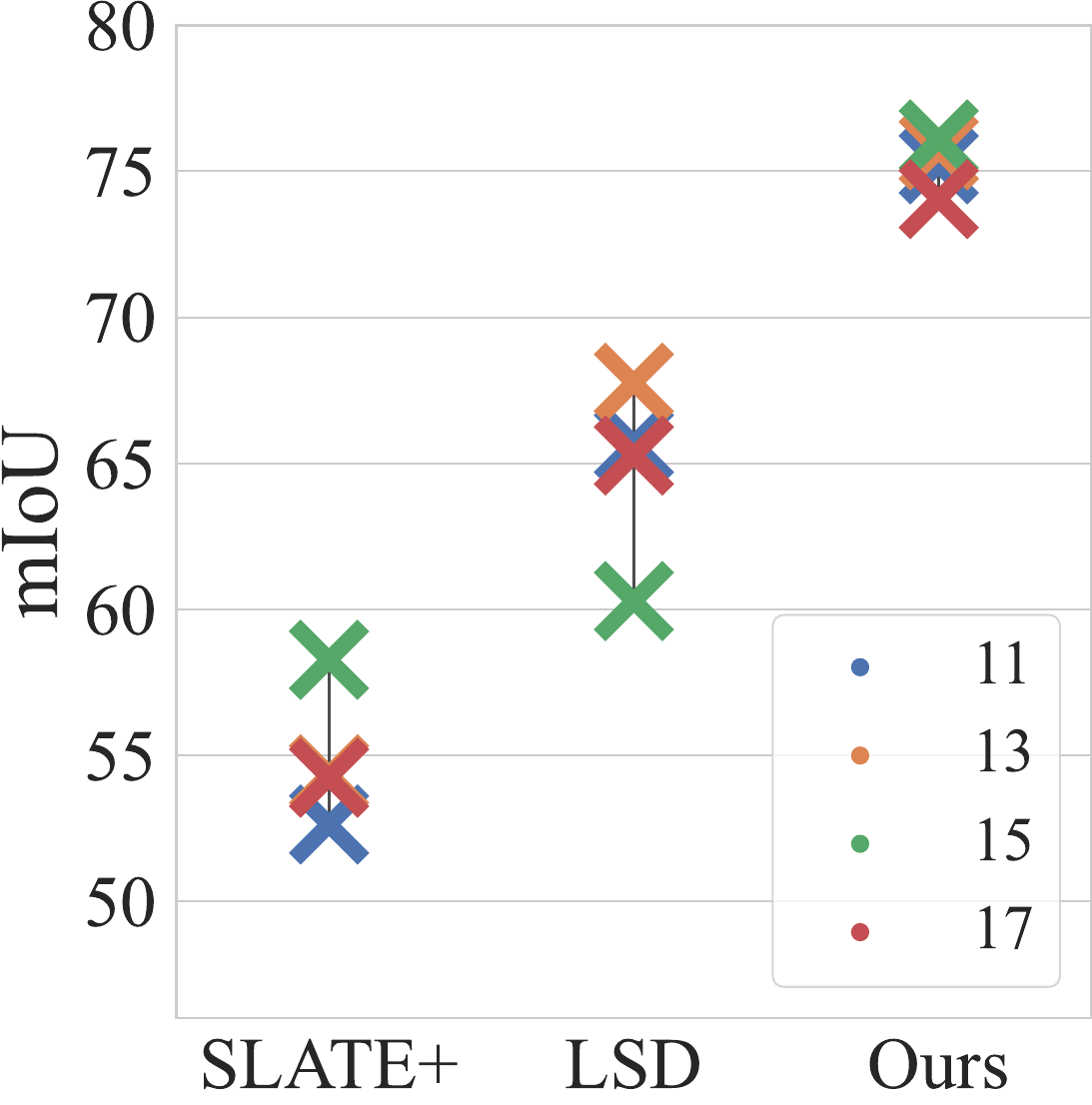}
    % \caption{Number of slots}
    % \label{subfig:robust_num_slot}
  \end{subfigure}
\hspace{0.2cm}
  \hfill
  \begin{subfigure}{0.3\textwidth} % Adjust the width as needed
    \centering
    \includegraphics[width=\linewidth]{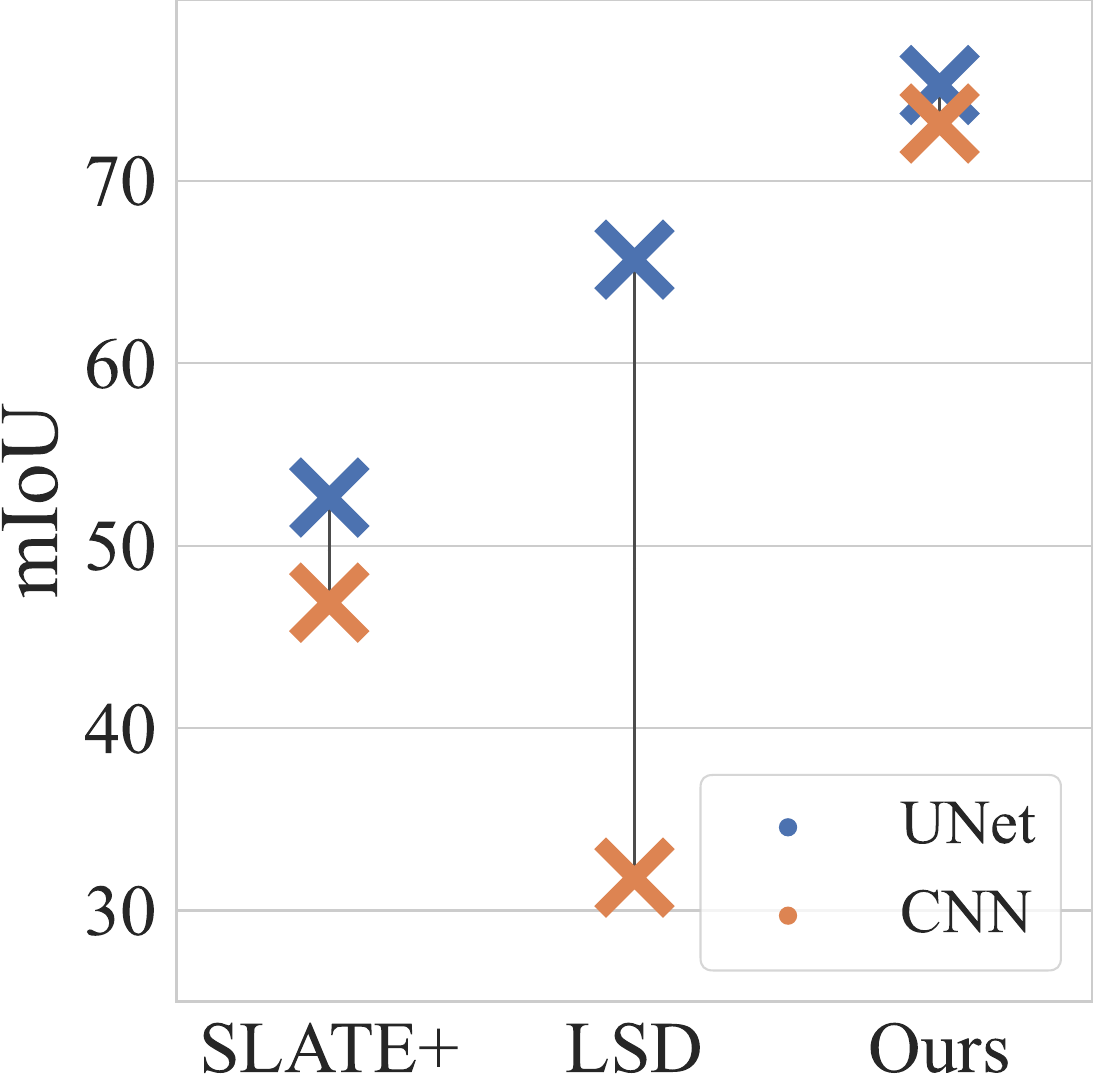}
    % \caption{Encoder architecture}
    % \label{subfig:robust_enc_type}
  \end{subfigure}
  \hspace{0.2cm}
  \hfill
  \begin{subfigure}{0.3\textwidth} % Adjust the width as needed
    \centering
    \includegraphics[width=\linewidth]{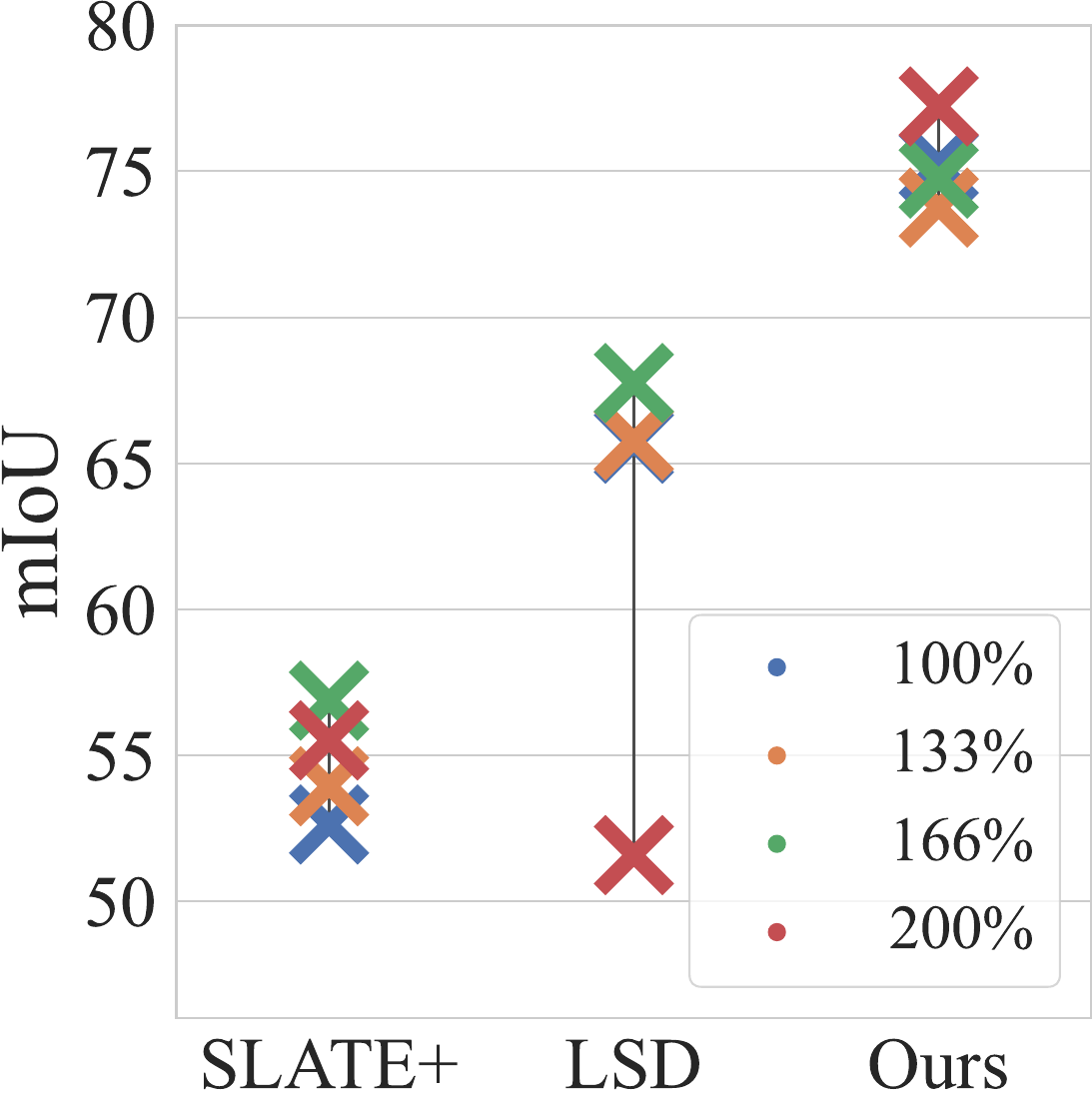}
    % \caption{Decoder capacity}
    % \label{subfig:robust_dec_cap}
  \end{subfigure}
  \begin{subfigure}{0.3\textwidth} % Adjust the width as needed
    \centering
    \includegraphics[width=\linewidth]{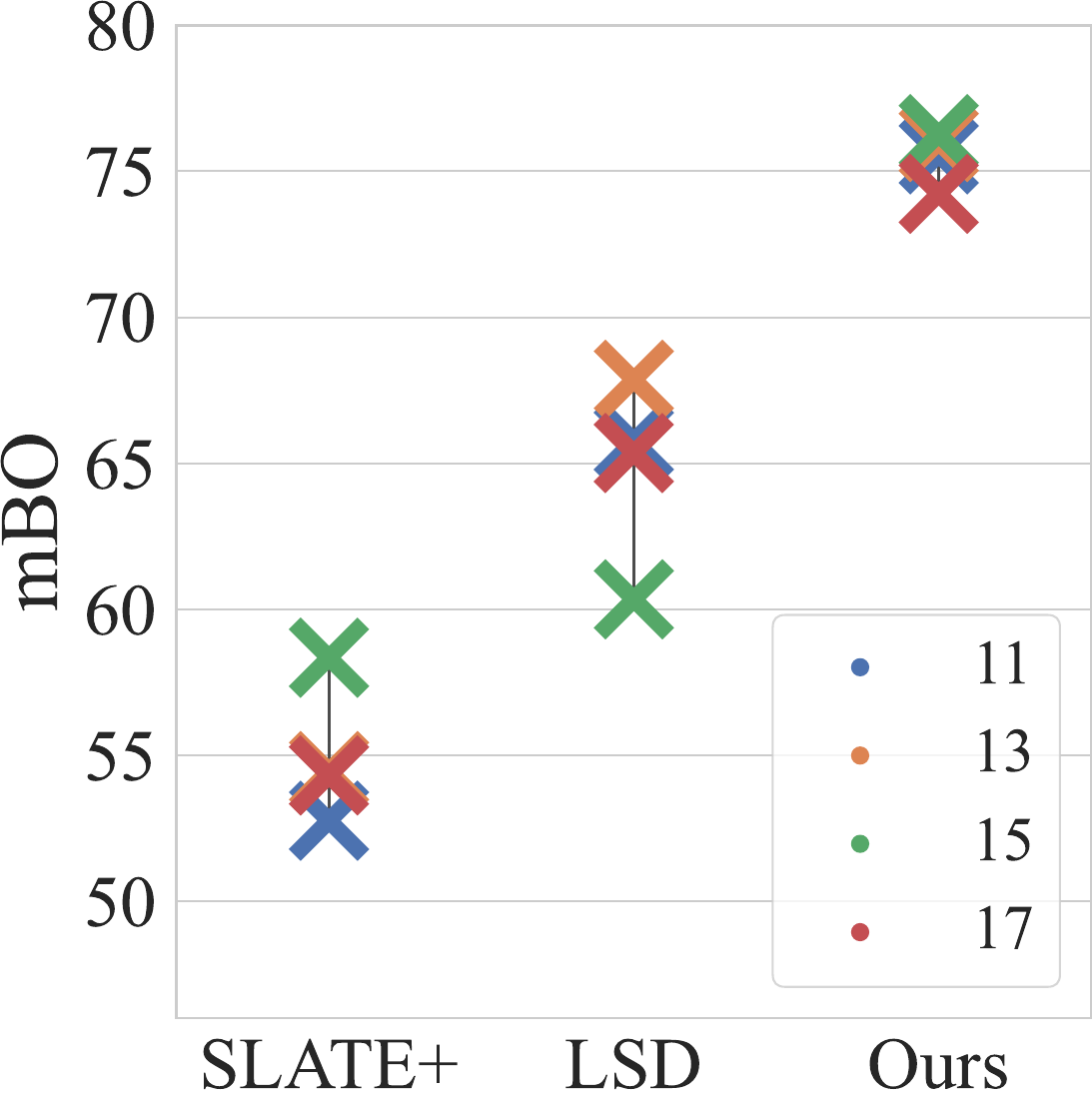}
    \caption{Number of slots}
    % \label{subfig:robust_num_slot}
  \end{subfigure}
\hspace{0.2cm}
  \hfill
  \begin{subfigure}{0.3\textwidth} % Adjust the width as needed
    \centering
    \includegraphics[width=\linewidth]{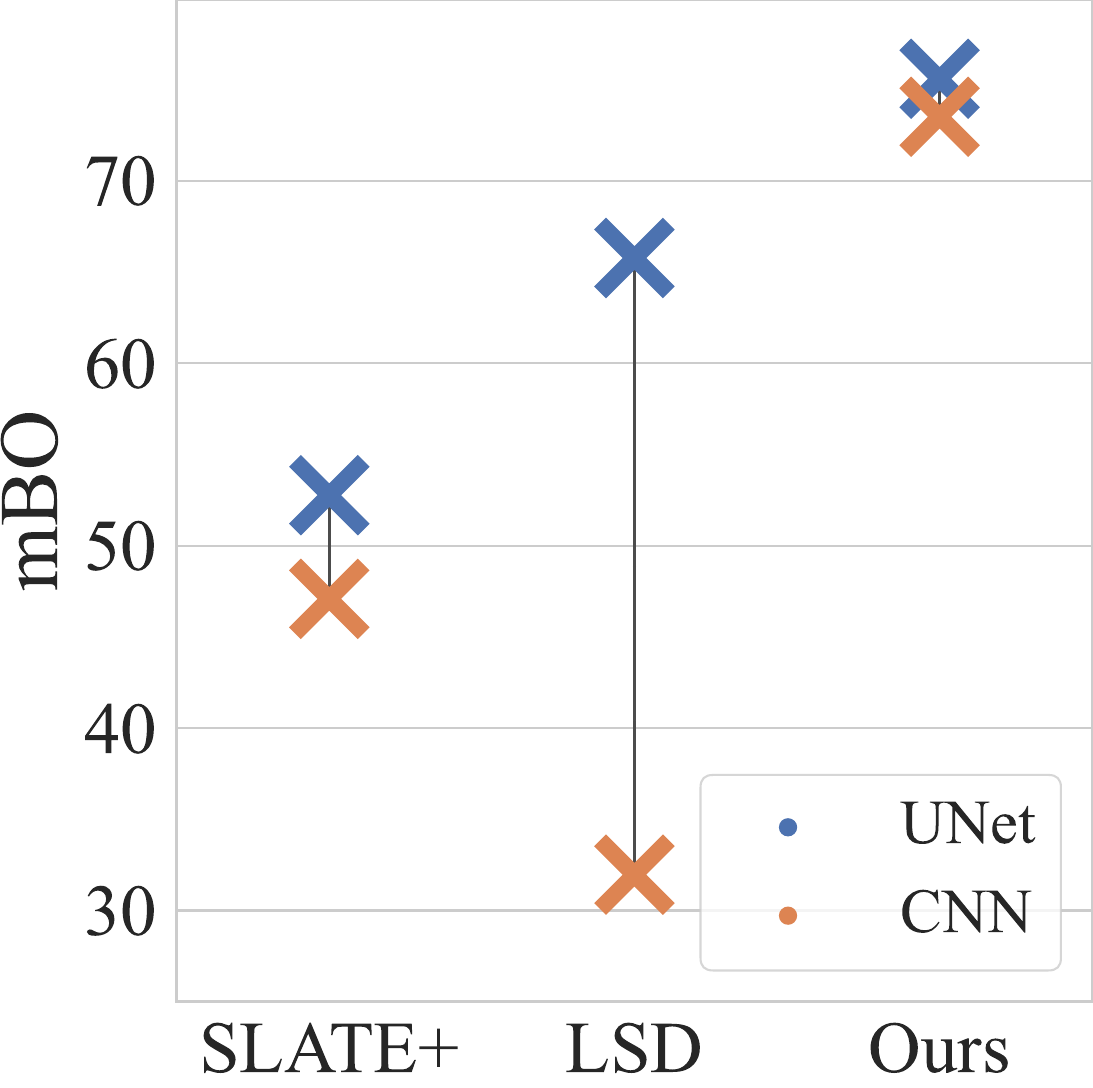}
    \caption{Encoder architecture}
    % \label{subfig:robust_enc_type}
  \end{subfigure}
  \hspace{0.2cm}
  \hfill
  \begin{subfigure}{0.3\textwidth} % Adjust the width as needed
    \centering
    \includegraphics[width=\linewidth]{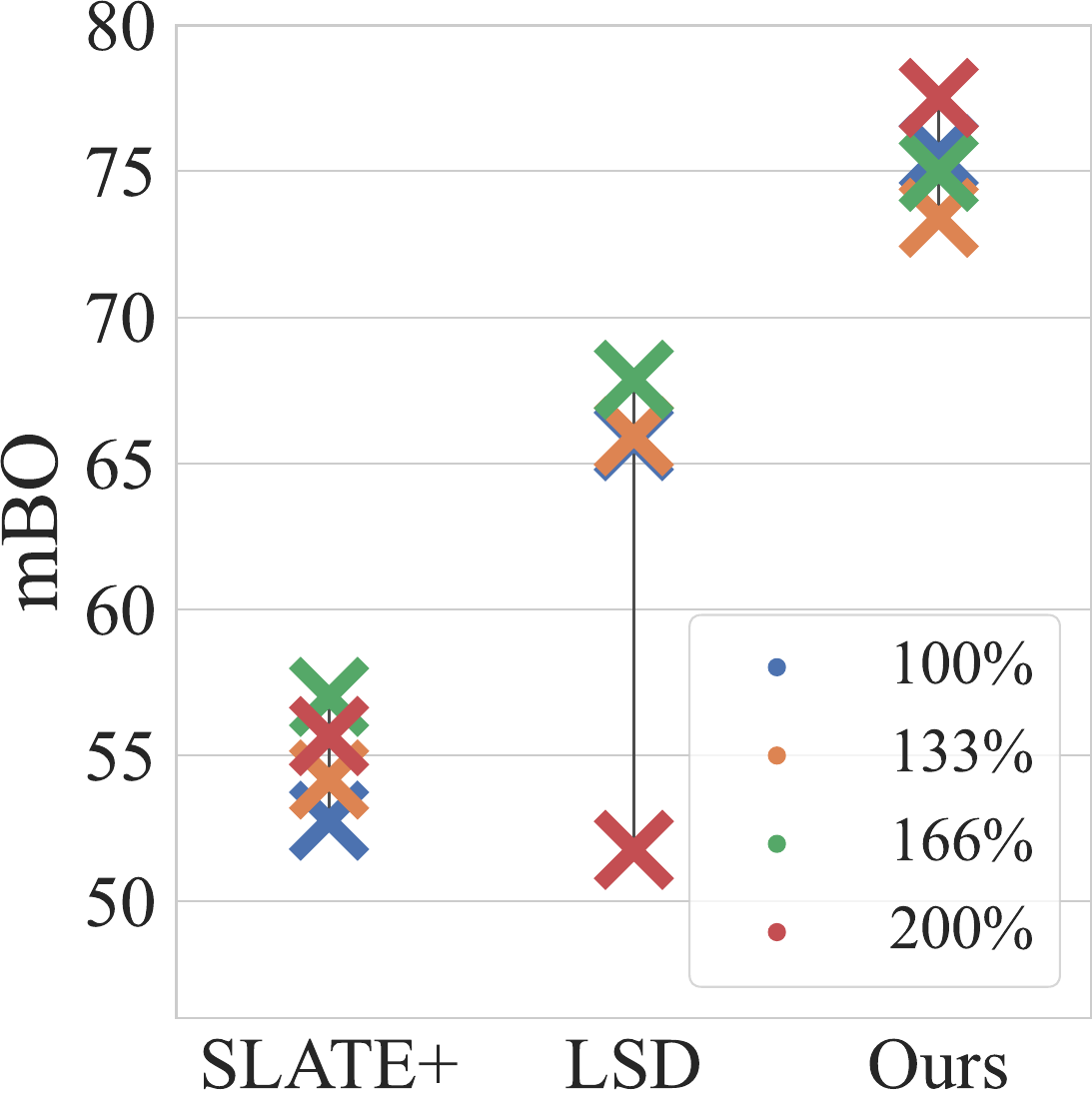}
    \caption{Decoder capacity}
    % \label{subfig:robust_dec_cap}
  \end{subfigure}
  \caption{\textbf{Robustness across Various Hyperparameters.} We evaluate the robustness of our model across different number of slots, encoder type, and decoder capacity. Among various hyperparameters, our model steadily shows powerful performance against baselines.}
  \label{fig:full_ablation_appendix}
  \vspace{-0.3cm}
\end{figure}

We include results of the robustness test on mIoU, mBO metrics in Figure~\ref{fig:full_ablation_appendix}. Similar to the results on FG-ARI (Figure~\ref{fig:robustness_all}), our model is surprisingly robust to a wide range of hyperparameters. 
It suggests that directly optimizing the compositionality of the representation significantly reduce a dependency on a choice of hyperparameters.

% \begin{figure}
% \centering
% \includegraphics[width=0.5\textwidth]{example-image-a}
% \caption{Full plots of ablation study on varying hyper-parameters}  
% \label{fig:full_ablation_appendix}
% \end{figure}

% \section{Additional Qualitative Results}  
\subsection{Unsupervised Object Segmentation}
We present additional qualitative results for unsupervised segmentation results in Figure~\ref{fig:unsupervised_seg_appendix}. 
Our method successfully segmented the object regions across all four datasets. 
In contrast, baselines easily divide each object into multiple segments or capture a wide area around the objects.

\subsection{Effect of Mixing Slot Strategy}
\begin{figure}
\centering
\includegraphics[width=1.0\textwidth]{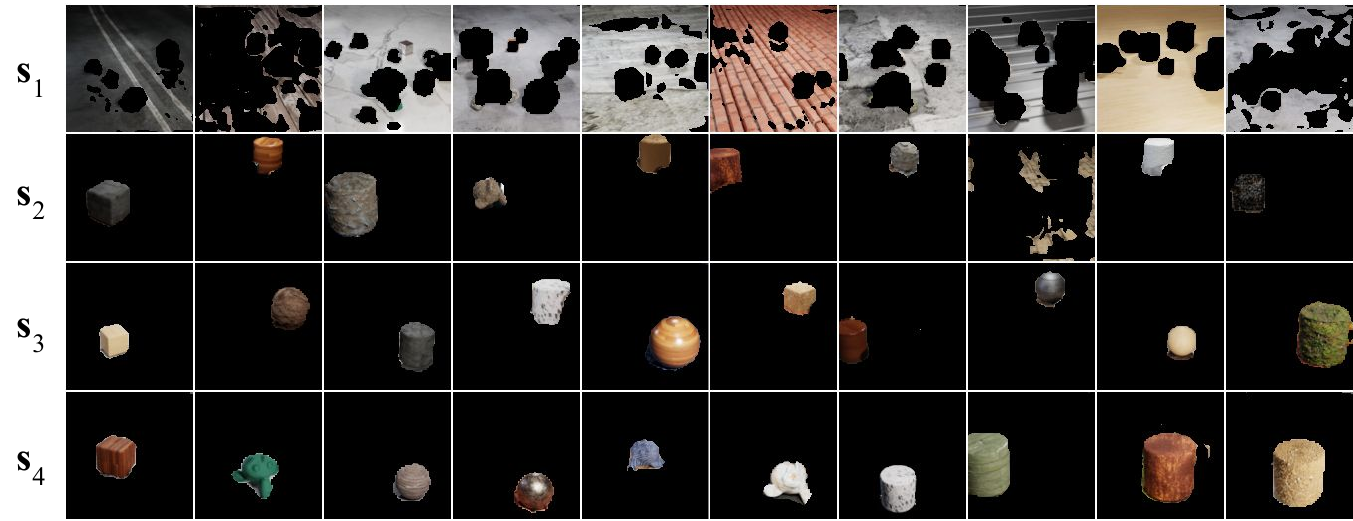}
\caption{\textbf{Grouping of slots by sharing the slot initialization.} We obtain slot representation from various images while sharing the initial values of $\mathbf{S}^{(0)}$ and cluster the representation based on their initial values. Slots initialized as $\mathbf{s}_1$ consistently capture backgrounds.}  
\label{fig:slot_init_share_appendix}
\end{figure}

\begin{figure}
\centering
\includegraphics[width=1.0\textwidth]{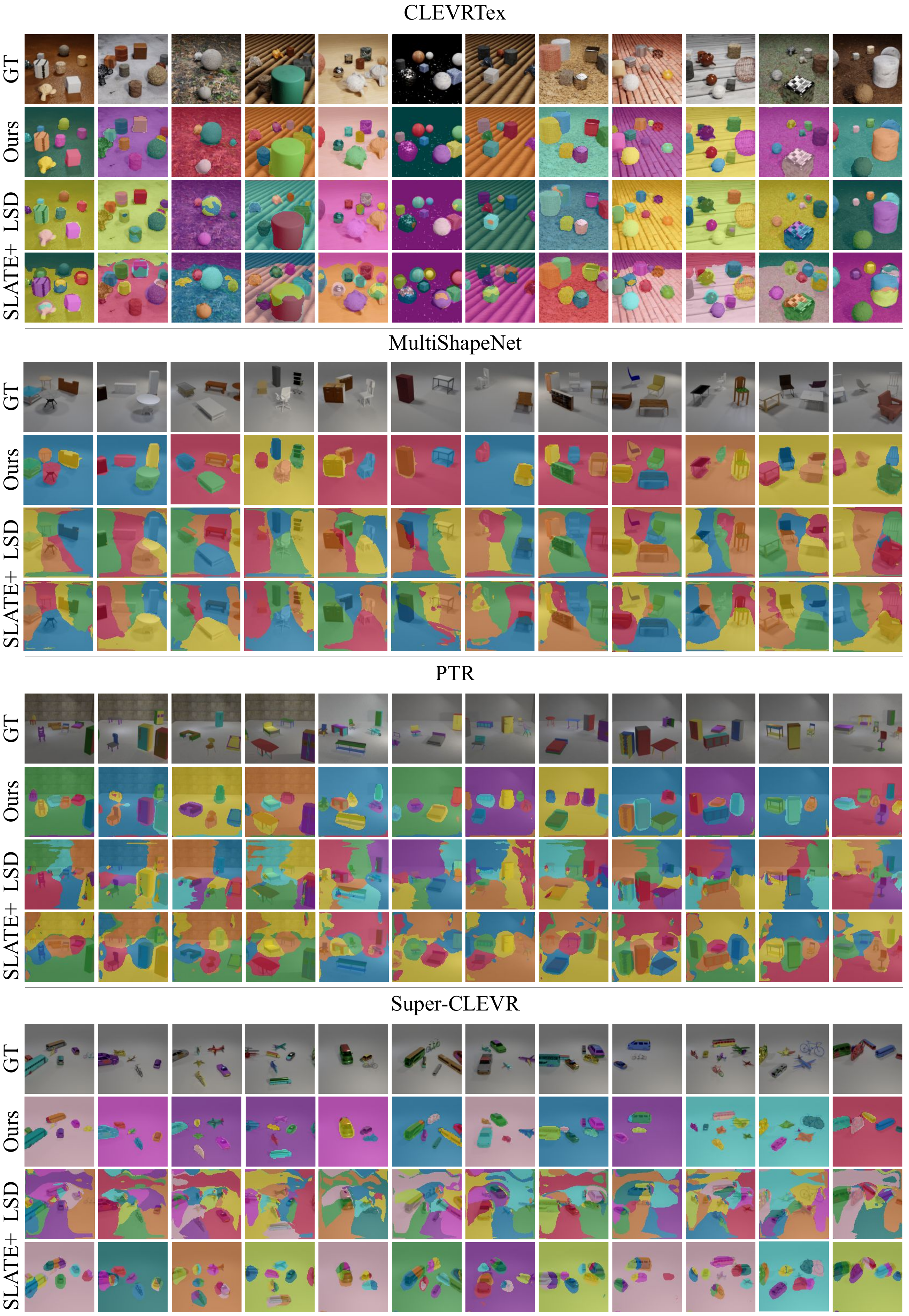}
\caption{\textbf{More Qualitative Results on Unsupervised Object Segmentation.}}  
\label{fig:unsupervised_seg_appendix}
\end{figure}

As discussed in Section~\ref{subsec:slot_mixing} and Section~\ref{subsec:ablation_study}, sharing $\mathbf{S}^{(0)}$ slightly enhances the performance by roughly avoiding suspicious compositions during training. 
To investigate how sharing slot initialization affects the composition, 
we obtained the slot representations from multiple scenes with the same slot initialization
and grouped those representations by their order, \textit{i.e.}, $\mathbf{s}_i$ belongs to $i$-th group.
Figure~\ref{fig:slot_init_share_appendix}, we observe that the captured objects from the same initialization are correlated to some degree. The slots in the first row mostly capture the backgrounds of the scenes, while other slots tend to capture foreground objects. Moreover, we observe that the slots in the fourth row tend to capture the objects located in the lower part of the scene. Based on these observations, we conjecture that sharing slot initialization stabilizes our framework by alleviating some suspicious compositions, such as the occlusion of foreground objects or composing multiple backgrounds. 
% In Section~\ref{subsec:slot_mixing}, we introduce two slot mixing strategies.

\begin{figure}
\centering
\includegraphics[width=1.0\textwidth]{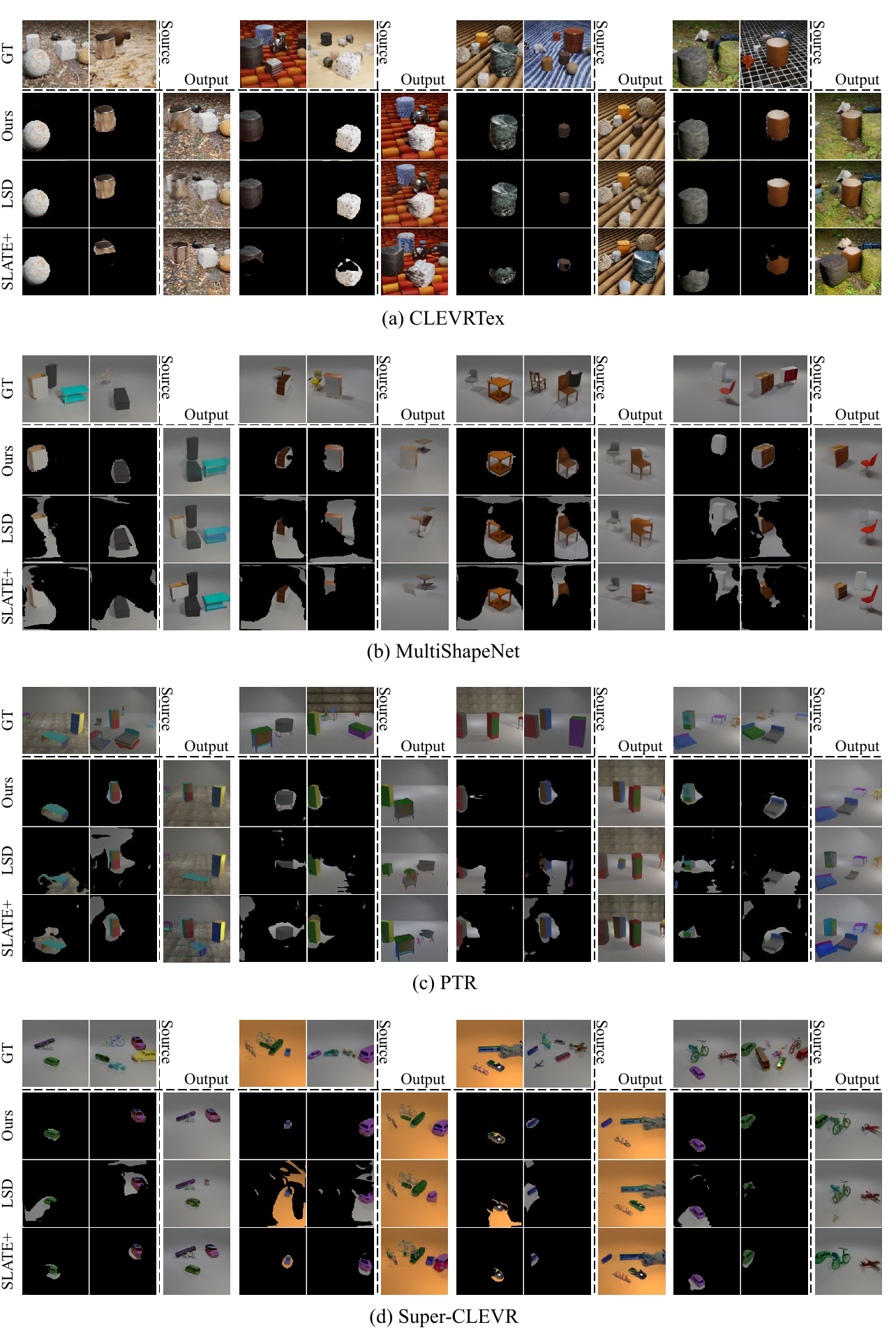}
\caption{\textbf{More Qualitative Results on Compositional Generation between two images.}}  
\label{fig:Composition_two_imgs_appendix}
\end{figure}

\subsection{Investigation on Compositionality of Slots}
\label{subsec:invest_compositionality}
In this section, we provide more visual samples of composite images to investigate the compositionality of slot representations in our method. 
Figure~\ref{fig:Composition_two_imgs_appendix} illustrates the results of generating composite images by mixing slots from two images, which supplements the Figure~\ref{fig:compositional_gen} in the main paper.
It shows that the baselines often fail to capture compositional objects into independent slots, while our method successfully learns object-level slots through the composition path.
As a result, the composite images generated by the baselines often fail to adhere the object-level manipulation, such as retaining the removed objects or transforming the object identity and background pattern while adding a new object. 
In contrast, our method preserves these semantics more precisely based on accurate object slots.
\subsection{Additional qualitative results on compositional generation}
\begin{figure}
\centering
\includegraphics[width=1.0\textwidth]{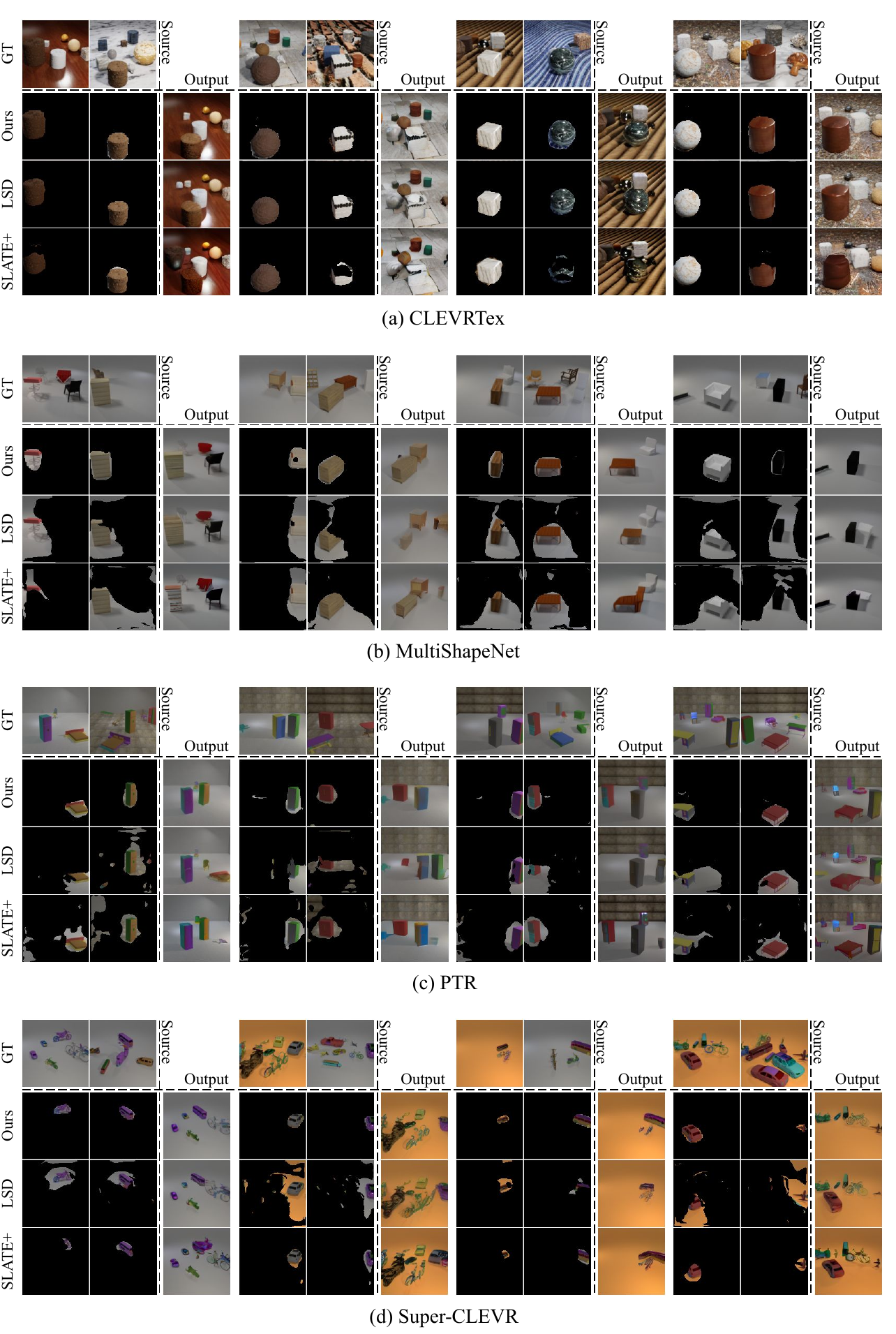}
\caption{\textbf{More Qualitative Results on Compositional Generation between two images.}}  
\label{fig:more_composition}
\end{figure}

To help a comprehensive understanding of the baselines, we provide more qualitative samples on compositional generation in Figure~\ref{fig:more_composition}. While Figure~\ref{fig:compositional_gen} and Figure~\ref{fig:Composition_two_imgs_appendix} illustrates the common failure cases of the baselines, we additionally present compositional generation results where the baselines also reasonably capture an object into a slot. Despite the reasonable slot attention masks, the composite image produced by the baseline model often distorts the original appearance of the object or creates unrealistic partial objects. In contrast, our model consistently produces faithful composite images, which highlights the importance of the compositional objective.

\clearpage

\subsection{Additional Evaluation on Object Property Prediction}
To assess the quality of acquired object representations, we employ object property prediction using the learned representation, following the methodology outlined in \cite{jiang23_lsd, dittadi2021generalization_and_robustness_ocl}.
During this process, we train a network to predict the property based on a fixed slot representation.
The true label for the slot representation is established through Hungarian matching, comparing the mask of slots with the foreground objects. The remaining slots after matching are considered as backgrounds.
For predicting properties, we employ a 4-layer MLPs with a hidden dimension of 196. Accuracy is reported for categorical properties, while mean squared error is reported for continuous properties. We assess the models on datasets that include object properties.

% overall table summary
%   our model --> consistently better
% specific description on each property
%   shape, position --> correlated to high segmentation quality.
%   material (texture) --> also can capture local? high-frequency? details effectively.
% explain the high MSE in Super-CLEVR position
%   compared to other datasets, lots of occlusion, small objects.
%   

The results for object property prediction are presented in Table~\ref{tab:downstream}. Our model consistently performs better than the baselines across different properties and datasets. Notably, it excels in predicting shape and position, as observed in the high segmentation performance depicted in Figure~\ref{fig:figure2_qual} and Table~\ref{tab:main}.
Furthermore, our model demonstrates improved performance in predicting materials indicating its ability to capture local and high-frequency information.

On the Super-CLEVR dataset, despite our model's higher segmentation performance, the mean square error of position remains competitive with other baselines. We attribute this to the challenging nature of the dataset, where scenes often include many small and occluded objects. As a result, both our model and the baselines face increased difficulty in predicting position, leading to a higher error rate compared to other datasets.

% \usepackage{graphicx}
\begin{table}[t]
\centering
\caption{\textbf{Results on object property prediction.} We evaluate the quality of the learned representation through object property prediction. Our model consistently performs better than the baselines across different properties and datasets.}
\label{tab:downstream}
\resizebox{0.95\textwidth}{!}{%
\begin{tabular}{c|ccc|cc|ccc}
\toprule
    Dataset & \multicolumn{3}{c|}{CLEVRTex} & \multicolumn{2}{|c|}{PTR} & \multicolumn{3}{|c}{Super-CLEVR} \\
    \midrule
    \multirow{2}{*}{Property} &  Position&  Shape&  Material&  Position&  Shape&  Position&  Shape& Material\\
            & $(\downarrow)$ & $(\uparrow)$ &  $(\uparrow)$ & $(\downarrow)$ &  $(\uparrow)$ & $(\downarrow)$ & $(\uparrow)$ &  $(\uparrow)$ \\
         \midrule
         SLATE+& 0.1757& 78.72& 67.99& 0.2218& 88.21& 0.5397& 76.28& 68.43\\
         LSD& 0.1563& 85.07& 82.33& 0.5999& 75.80& 0.4372& 76.5& 69.24\\
         Ours& \textbf{0.1044}& \textbf{88.86}& \textbf{84.29}& \textbf{0.1424}& \textbf{90.00}& \textbf{0.4262}& \textbf{80.67}& \textbf{71.31}\\
     \bottomrule
\end{tabular}%
}
\end{table}

\clearpage
\subsection{Additional Results on Real-world dataset}

To explore the scalability of our novel objective in a complex real-world dataset, we examine our framework in BDD100k dataset~\cite{bdd100k}, which consists of diverse driving scenes.   Since the images captured on night or rainy days often produce blurry and dark images, we filter the data to collect only sunny and daytime images using metadata, which leaves about 12k, 1.7k images in the training/validation set, respectively. 
% Since training in complex real-world datasets usually requires magnificent time and data, we bootstrap our auto-encoding path with off-the-shelf models following \cite{jiang23_lsd}. Specifically, we employ pretrained DINOv2~\cite{oquab2023dinov2}, 
Since it has been widely observed that learning the object-centric representation directly on real-world dataset is challenging, we bootstrap our auto-encoding path with off-the-shelf models following \cite{jiang23_lsd}. Specifically, we employ pretrained DINOv2~\cite{oquab2023dinov2} and Stable Diffusion~\cite{rombach22_ldm} for the image encoder and slot decoder in our auto-encoding path, respectively. 
Instead of using frozen Stable-Diffusion, we update key and value mapping layers in cross-attention layers to enhance the overall auto-encoding performance following \cite{kumari23_custom_diffusion}. 
% For efficient training, we first warm up the auto-encoding path with only auto-encoding objective for 200k iterations and then train only the surrogate decoder for 140k iterations on top of frozen slot representations, which found to significantly boost up the convergence of the surrogate decoder. 
For efficient training, we first warm up the auto-encoding path for 200k iterations and then train only the surrogate decoder for 140k iterations on top of frozen slot representations, which significantly boosts up the convergence of the surrogate decoder. 
Finally, we optimize our compositional path for 100k iterations. 
% For the baseline, we compare our model training only with auto-encoding objective for 300k iterations. \cite{jiang23_lsd} also introduces this auto-encoding framework and named Stable-LSD. 
For the baseline, we compare our model trained with only auto-encoding objective for 300k iterations, which converges closely to the Stable-LSD \cite{jiang23_lsd}. 

\begin{figure}
\centering
\includegraphics[width=1.0\textwidth]{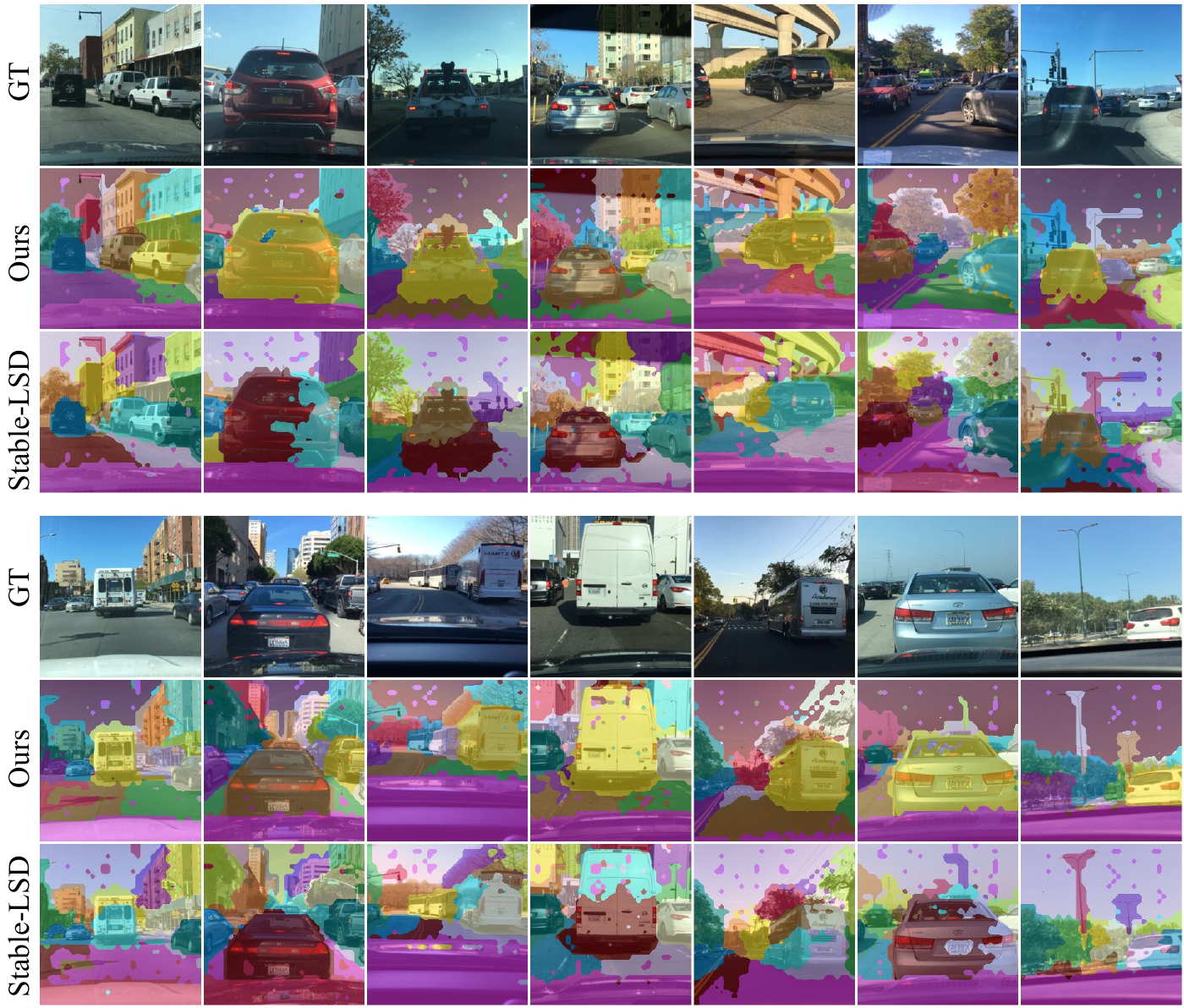}
\caption{\textbf{Qualitative results on unsupervised segmentation in BDD100k.}}  
\label{fig:bdd100k_segmentation}
\end{figure}

Figure~\ref{fig:bdd100k_segmentation} illustrates qualitative results on unsupervised object segmentation. The slot attention masks of our model successfully capture composable instances such as cars, buildings, trees, font hoods, etc. In contrast, the diffusion model trained without compositional objective often divides the objects into multiple slots or encodes multiple objects into a slot. For example, the car or truck is frequently divided into multiple masks, and multiple cars are often encoded into a single slot. 

\begin{figure}
\centering
\includegraphics[width=1.0\textwidth]{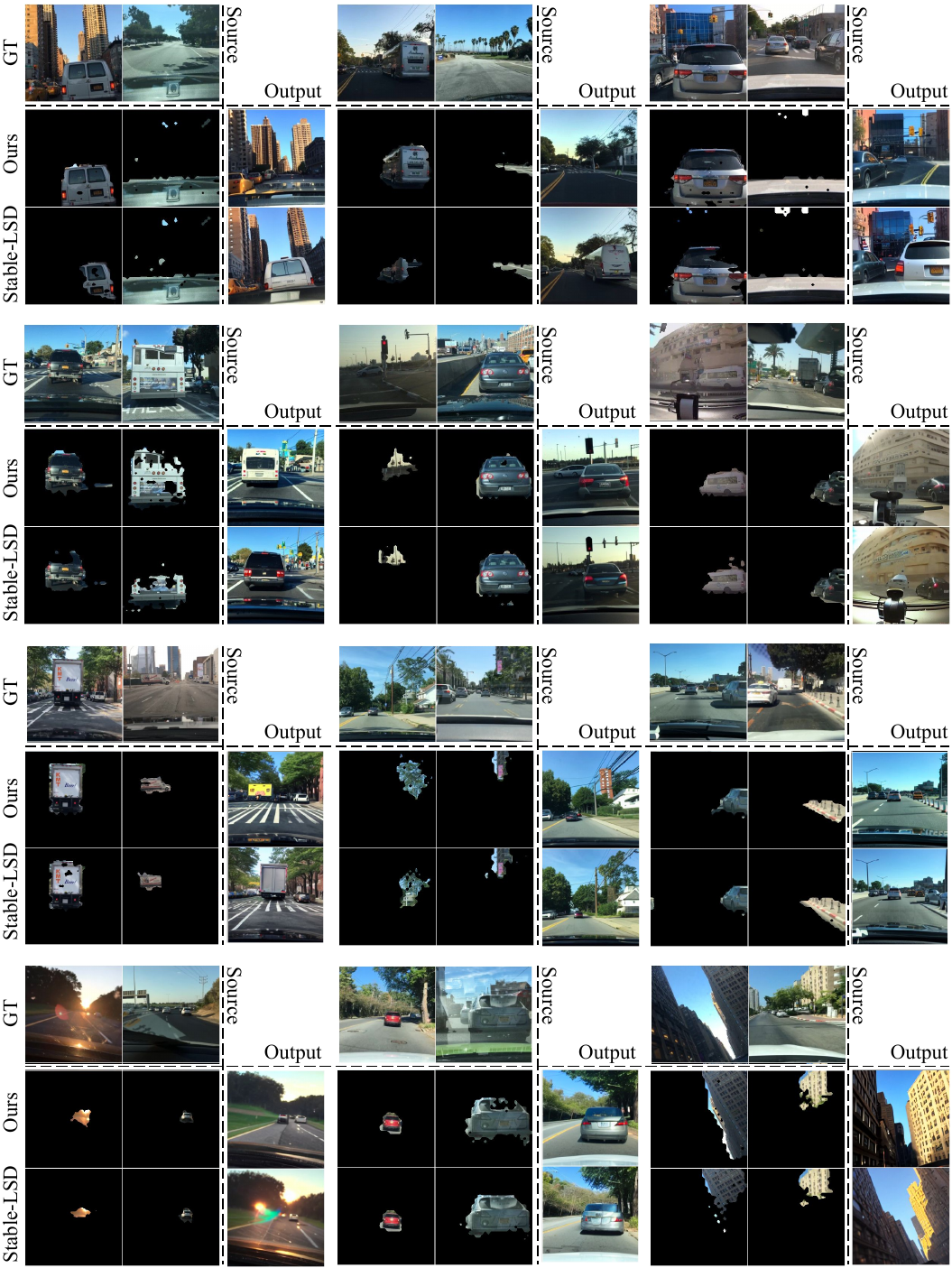}
\caption{\textbf{Qualitative results on compositional generation in BDD100k.}}  
\label{fig:bdd100k_composition}
\end{figure}

To further examine the compositionality of the learned slot representations, we qualitatively analyze the visual samples of composite images in Figure~\ref{fig:bdd100k_composition} similar to Section~\ref{subsec:invest_compositionality}. 
% We identify that our model successfully generates realistic samples even in complex and diverse scenes thanks to the compositional objective. 
We observe that our method successfully generates realistic scenes, modeling complex correlations among objects and environments. It appropriately adapts the appearance of newly added/removed objects, their shadow, reflections in the front glass and hood, and sometimes even global illumination change caused by removing the sun.
In contrast, the auto-encoding model often fails to achieve faithful composition. For example, in Row 1 of Figure~\ref{fig:bdd100k_composition}, the car still appears in the composite image even after the removal of the corresponding slot. 
Also, we observe that removing slots containing partial information of the object often leads to undesirable artifacts in composite images such as creating a new car in the first example of Row 2, or leaving unrealistic artifacts in the third example of Row 2. In contrast, our model produces natural object-wise manipulation. Moreover, the baseline model often fails to faithfully generate the inserted object as shown in Row 3, while our model tends to maintain the target object. In Row 4, we identify that our model successfully models complex interaction between slots such as removing sunlight changing the reflection of the bonnet in the first image, or changing a blurry car into a sharp car corresponding to bright weather. In summary, we identify that our novel objective on compositionality can help to learn object-wise disentanglement even in complex scenes and helps to model complex interactions among objects.

\end{document}